%% file: nips_2016.tex
\documentclass{article}

\PassOptionsToPackage{numbers, compress}{natbib}
\usepackage[final]{nips_2016}

\usepackage[utf8]{inputenc} 
\usepackage[T1]{fontenc}    
\usepackage{url}            
\usepackage{booktabs}       
\usepackage{amsfonts}       
\usepackage{hyperref}
\usepackage{amsmath}        
\usepackage{nicefrac}       
\usepackage{graphicx}       
\usepackage{microtype}      
\usepackage{subfig}      
\usepackage{amsmath}
\usepackage[font={footnotesize}]{caption}
\usepackage{enumitem}

\usepackage{rotating}
\usepackage{array}

\setlist{nosep}

\usepackage{color}

\title{Progressive Neural Networks}

\author{Andrei A. Rusu*, \ Neil C. Rabinowitz*, \ Guillaume Desjardins*, \ Hubert Soyer,\\
\textbf{James Kirkpatrick, \ Koray Kavukcuoglu, \ Razvan Pascanu, \ Raia Hadsell} \\
\small{* These authors contributed equally to this work}
\vspace{0.15 cm}
\\
Google DeepMind \\
London, UK
\vspace{0.15 cm}
\\
\small\texttt{\{andreirusu, ncr, gdesjardins, soyer, kirkpatrick, korayk, razp, raia\}@google.com} \\
}
\begin{document}

\maketitle

\begin{abstract}
Learning to solve complex sequences of tasks---while both leveraging transfer and
avoiding catastrophic forgetting---remains a key obstacle to achieving human-level
intelligence. The \emph{progressive networks} approach represents a step forward in this direction:
they are immune to forgetting and can leverage prior knowledge via lateral
connections to previously learned features. We evaluate this architecture
extensively on a wide variety of reinforcement learning tasks (Atari and 3D
maze games), and show that it outperforms common baselines based on pretraining
and finetuning. Using a novel sensitivity measure, we demonstrate
that transfer occurs at both low-level sensory and high-level control layers of the learned policy.
\end{abstract}

\input{intro}
\input{progressive}
\input{method_transfer}
\input{related_work}
\input{experiment_setup}
\input{infinite_pong}

\input{two_columns}

\input{atari_analysis}

\input{labyrinth}
\input{discussion}

\newpage
\footnotesize
\setlength{\bibsep}{5pt}
\bibliographystyle{plain}
\bibliography{progressive}

\newpage

\appendix
\pagenumbering{arabic}

{\Large{\textbf{Supplementary Material}}}

\input{app_perturbation}
\input{app_compression}

\input{app_setup}
\input{app_curves}

\input{app_lab}

\end{document}

%% file: intro.tex
\section{Introduction}

Finetuning remains the method of choice for transfer learning with neural
networks: a model is pretrained on a source domain (where data is often
abundant), the output layers of the model are adapted to the target domain, and the
network is finetuned via backpropagation. This approach was
pioneered in \cite{hinton2006science} by transferring knowledge from a
generative to a discriminative model, and has since been generalized
with great success \cite{UTLC+LISA-2011}.  Unfortunately, the approach has drawbacks
which make it unsuitable for transferring across \textit{multiple tasks}: if we
wish to leverage knowledge acquired over a sequence of experiences, which model
should we use to initialize subsequent models? This seems to require not only a
learning method that can support transfer learning without catastrophic forgetting, but also
foreknowledge of task similarity. Furthermore, while finetuning may allow us to
recover expert performance in the target domain, it is a destructive process
which discards the previously learned function. One could copy each model before
finetuning to explicitly remember all previous tasks, but the issue of
selecting a proper initialization remains.  While distillation
\cite{Hinton2015distil} offers one potential solution to multitask learning
\cite{Rusu15}, it requires a reservoir of persistent training data for all tasks,
an assumption which may not always hold.

This paper introduces \textit{progressive networks}, a novel model architecture
with explicit support for transfer across sequences of tasks. While finetuning
incorporates prior knowledge only at initialization, progressive networks
retain a pool of pretrained models throughout training, and learn lateral connections
from these to extract useful features for the new task.  By combining previously
learned features in this
manner, progressive networks achieve a richer compositionality, in which prior
knowledge is no longer transient and can be integrated at each layer of the
feature hierarchy. Moreover, the addition of new capacity alongside pretrained
networks gives these models the flexibility to both reuse old computations and learn new ones.
As we will show, progressive networks naturally accumulate experiences and
are immune to catastrophic forgetting by design, making them an ideal
springboard for tackling long-standing problems of continual or lifelong
learning.

The contributions of this paper are threefold. While many of the individual
ingredients used in progressive nets can be found in the literature, their
combination and use in solving complex sequences of tasks is novel. Second,
we extensively evaluate the model in complex reinforcement learning domains.
In the process, we also evaluate alternative approaches to transfer
(such as finetuning) within the RL domain. In particular, we show that
progressive networks provide comparable (if not slightly better) transfer
performance to traditional finetuning, but without the destructive
consequences. Finally, we develop a novel analysis based on
Fisher Information and perturbation which allows us to analyse in
detail how and where transfer occurs across tasks.

%% file: progressive.tex
\section{Progressive Networks}

Continual learning is a long-standing goal of machine learning, where agents
not only learn (and remember) a series of tasks experienced in sequence, but also have
the ability to transfer knowledge from previous tasks to improve convergence speed \cite{AAAIMag11-Taylor}.
\textit{Progressive networks} integrate these desiderata directly into the
model architecture: catastrophic forgetting is prevented by instantiating a new
neural network (a \textit{column}) for each task being solved, while transfer is
enabled via lateral connections to features of previously learned columns. The
scalability of this approach is addressed at the end of this section.

\begin{figure}[h]
  \centering
    \includegraphics[width=.25\textwidth]{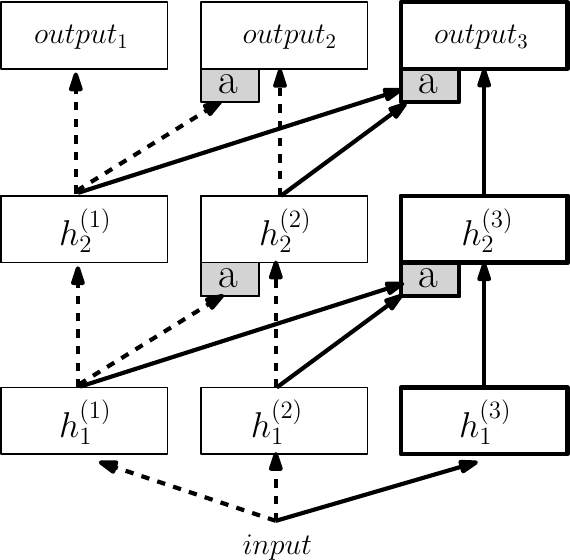}
    \caption{Depiction of a three column progressive network. The first two
columns on the left (dashed arrows) were trained on task 1 and 2 respectively.
The grey box labelled $a$ represent the adapter layers (see text).
A third column is added for the final task having access to all previously learned
features.
    }
    \label{fig:progressiveNet}
\end{figure}

A progressive network starts with a single column: a deep neural network having
$L$ layers with hidden activations $h_i^{(1)} \in \mathbb{R}^{n_i}$,
with $n_i$ the number of units at layer $i \le L$, and parameters
$\Theta^{(1)}$ trained to convergence.  When switching to a second task, the
parameters $\Theta^{(1)}$ are ``frozen'' and a new
column with parameters $\Theta^{(2)}$ is instantiated (with random
initialization), where layer $h_i^{(2)}$ receives input from both
$h_{i-1}^{(2)}$ and $h_{i-1}^{(1)}$ via lateral connections. This
generalizes to $K$ tasks as follows:
\footnote{Progressive networks can also be generalized in a straightforward
manner to have arbitrary network width per column/layer, to accommodate varying
degrees of task difficulty, or to compile lateral connections from multiple,
independent networks in an ensemble setting. Biases are omitted for clarity.}:
\begin{align}
  \label{eq:prognet}
  h_i^{(k)} = f\left( W_i^{(k)} h_{i-1}^{(k)} + \sum_{j<k} U_{i}^{(k:j)} h_{i-1}^{(j)} \right),
\end{align}
where $W_i^{(k)} \in \mathbb{R}^{n_{i} \times n_{i-1}}$ is the weight matrix of layer
$i$ of column $k$, $U_{i}^{(k:j)} \in \mathbb{R}^{n_i \times n_j}$ are the
lateral connections from layer $i-1$ of column $j$, to layer $i$ of column $k$
and $h_0$ is the network input.
$f$ is an element-wise non-linearity: we use $f(x)=\max(0, x)$ for all
intermediate layers.
A progressive network with $K=3$ is shown in Figure~\ref{fig:progressiveNet}.

These modelling decisions are informed by our desire to:
(1) solve $K$ independent tasks at the end of training;
(2) accelerate learning via transfer when possible; and
(3) avoid catastrophic forgetting.

In the standard pretrain-and-finetune paradigm, there is often an implicit
assumption of ``overlap'' between the tasks. Finetuning is efficient in this
setting, as parameters need only be adjusted slightly to
the target domain, and often only the top layer is retrained \cite{yosinski-nips2014}. In contrast, we make no assumptions about the
relationship between tasks, which may in practice be orthogonal or even
adversarial. While the finetuning stage could potentially unlearn these
features, this may prove difficult. Progressive networks side-step
this issue by allocating a new column for each new task, whose weights are
initialized randomly. Compared to the task-relevant initialization of pretraining,
columns in progressive networks are free to reuse, modify or ignore previously
learned features via the lateral connections.
As the lateral connections $U_{i}^{(k:j)}$ are only from column $k$ to columns
$j < k$, previous columns are not affected by the newly learned features in the
forward pass.  Because also the parameters $\{ \Theta^{(j)}; j<k\}$
are kept frozen (i.e. are constants for the optimizer) when training $\Theta^{(k)}$,
there is no interference between
tasks and hence no catastrophic forgetting.

\paragraph{Application to Reinforcement Learning.} Although progressive
networks are widely applicable, this paper focuses on their application to
deep reinforcement learning. In this case, each column is trained to solve a
particular Markov Decision Process (MDP): the $k$-th column thus defines a policy
$\pi^{(k)}(a\mid s)$ taking as input a state $s$ given by the environment,
and generating probabilities over actions $\pi^{(k)}(a\mid s) := h_L^{(k)}(s)$.
At each time-step, an action is sampled from this distribution and taken in the
environment, yielding the subsequent state. This policy implicitly defines a
stationary distribution $\rho_{\pi^{(k)}}(s,a)$ over states and actions.

\paragraph{Adapters.}
In practice, we augment the progressive network layer of Equation~\ref{eq:prognet} with
non-linear lateral connections which we call \textit{adapters}. They serve
both to improve initial conditioning and perform dimensionality reduction.
Defining the vector of anterior features
$h_{i-1}^{(<k)} = [h_{i-1}^{(1)} \cdots h_{i-1}^{(j)} \cdots h_{i-1}^{(k-1)}]$
of dimensionality $n_{i-1}^{(<k)}$, in the case of dense layers,
we replace the linear lateral connection with a single hidden layer MLP.
Before feeding the lateral activations into the MLP, we multiply them by a learned scalar,
initialized by a random small value.
Its role is to adjust for the different scales of the different inputs.
The hidden layer of the non-linear adapter is a projection onto an $n_{i}$ dimensional
subspace.  As the
index $k$ grows, this ensures that the number of parameters stemming from the
lateral connections is in the same order as $\left|\Theta^{(1)} \right|$. Omitting bias
terms, we get:
\begin{align}
  \label{eq:prognet}
  h_i^{(k)} = \sigma \left( W_i^{(k)} h_{i-1}^{(k)} + U_{i}^{(k:j)} \sigma(V_{i}^{(k:j)} \alpha_{i-1}^{(<k)} h_{i-1}^{(<k)}) \right),
\end{align}
where $V_{i}^{(k:j)} \in \mathbb{R}^{n_{i-1} \times n_{i-1}^{(<k)}}$ is the projection
matrix. For convolutional layers, dimensionality reduction is
performed via $1\times 1$ convolutions \cite{LinCY13}.

\paragraph{Limitations.}
\textit{Progressive networks} are a stepping stone towards a full continual
learning agent: they contain the necessary ingredients to learn multiple tasks,
in sequence, while enabling transfer and being immune to catastrophic
forgetting. A downside of the approach is the growth in number of
parameters with the number of tasks.
The analysis of Appendix 2 reveals
that only a fraction of the new capacity is actually utilized, and that this trend increases with more columns. This
suggests that growth can be addressed, e.g. by adding fewer layers or less capacity, by pruning \cite{Cun90optimalbrain}, or by online compression
\cite{Rusu15} during learning.  Furthermore, while progressive networks retain the
ability to solve all $K$ tasks at test time, choosing which column to use for
inference requires knowledge of the task label. These issues are left as future
work.

%% file: method_transfer.tex
\section{Transfer Analysis}
\label{sec_transfer}

Unlike finetuning, progressive nets do not destroy the features learned on
prior tasks. This enables us to study in detail which features and at which
depth transfer actually occurs. We explored two related methods: an intuitive,
but slow method based on a perturbation analysis, and a faster analytical method
derived from the Fisher Information \cite{amari98natural}.

\paragraph{Average Perturbation Sensitivity (APS).} To evaluate the degree to
which source columns contribute to the target task, we can inject
Gaussian noise at isolated points in the architecture (e.g. a given
layer of a single column) and measure the impact of this perturbation
on performance. A significant drop in performance indicates that the
final prediction is heavily reliant on the feature map or
layer. We find that this method yields similar results to the faster
Fisher-based method presented below. We thus relegate details and
results of the perturbation analysis to the appendix.

\paragraph{Average Fisher Sensitivity (AFS).}
We can get a local approximation to the perturbation sensitivity by using
the Fisher Information matrix
\cite{amari98natural}. While the Fisher matrix is typically computed with
respect to the model
parameters, we compute a modified diagonal Fisher $\hat{F}$ of the network policy $\pi$
with respect to the \textit{normalized activations}
\footnote{The Fisher of individual neurons (fully connected) and feature maps
(convolutional layers) are computed over $\rho_{\pi^{(k)}}(s,a)$. The use of a normalized
representation $\hat{h}$ is non-standard, but makes the scale of $\hat{F}$
comparable across layers and columns.}
at each layer $\hat{h}_i^{(k)}$. For convolutional layers, we define
$\hat{F}$ to implicitly perform a summation over pixel locations. $\hat{F}$ can
be interpreted as the sensitivity of the policy to small changes in the
representation. We define the diagonal matrix $\hat{F}$, having
elements $\hat{F}(m,m)$, and the derived Average Fisher
Sensitivity (AFS) of feature $m$ in layer $i$ of column $k$ as:
\begin{align*}
\hat{F}_i^{(k)} &=
     \mathbb{E}_{\rho(s, a)}
        \left[
          \frac {\partial \log \pi} {\partial \hat{h}_i^{(k)}} \,
          \frac {\partial \log \pi} {\partial \hat{h}_i^{(k)}}^T
        \right]
& \hspace{1cm} &
\text{AFS}(i,k,m) &=
    \frac{\hat{F}_i^{(k)}(m,m)}{\sum_k \hat{F}_i^{(k)}(m,m)}
\end{align*}
where the expectation is over the joint state-action distribution $\rho(s,a)$
induced by the progressive network
trained on the target task. In practice, it is often useful to consider the AFS
score per-layer $\text{AFS}(i,k) = \sum_m \text{AFS}(i,k,m)$, i.e. summing over
all features of layer $i$. The AFS and APS thus estimate how much
the network relies on each feature or column in a layer to compute its output.

%% file: related_work.tex
\section{Related Literature}

There exist many different paradigms for transfer and
multi-task reinforcement learning, as these have long been recognized 
as critical challenges in AI research
\cite{Ring1995,SilverYL13,AAAIMag11-Taylor}. Many methods for transfer
learning rely on linear and other simple models
(e.g. \citep{Ruvolo2013ELLA}), which is a limiting factor to their
applicability. Recently, there have been new methods proposed for
multi-task or transfer learning with deep RL:
\cite{tessler2016,Rusu15,ParisottoICLR16}. In this
work we present an architecture for deep reinforcement learning that
in sequential task regimes that enables learning without forgetting
while supporting individual feature transfer from previous
learned tasks.

Pretraining and finetuning was proposed in \citep{hinton2006science}
and applied to transfer learning in
\citep{Bengio12deeplearning,UTLC+LISA-2011}, generally
in unsupervised-to-supervised or supervised-to-supervised
settings. The actor-mimic approach
\cite{ParisottoICLR16} applied these principles to
reinforcement learning, by fine-tuning a DQN multi-task network on new
Atari games and showing that some responded with faster learning,
while others did not. Progressive networks differ from the
finetuning direction substantially, since capacity is added as new tasks are learned.

Progressive nets are related to the incremental and constructive
architectures proposed in neural network literature. The
cascade-correlation architecture was designed to eliminate forgetting
while incrementally adding and refining feature extractors
\cite{Fahlman1990}.  Auto-encoders such as \citep{ZhouAISTATS12} use
incremental feature augmentation to track concept drift, and deep
architectures such as \citep{RozantsevSF16} have been designed that
specifically support feature transfer.  More recently, in
\cite{Agostinelli_NIPS2013}, columns are separately trained on
individual noise types, then linearly combined, and 
\cite{CiresanMS12} use columns for image classification. The block-modular architecture of \cite{Terekhov2015} has many similarities to our approach but focuses on a visual discrimination task.
The progressive net approach, in contrast, uses lateral connections to access previously learned features for deep compositionality. It can be used in any sequential learning setting but is especially valuable in RL.

%% file: experiment_setup.tex
\section{Experiments}

\begin{figure}[h]
\centering
\subfloat[Pong variants]{
\begin{tabular}{cc}
  \includegraphics[width=.09\textwidth]{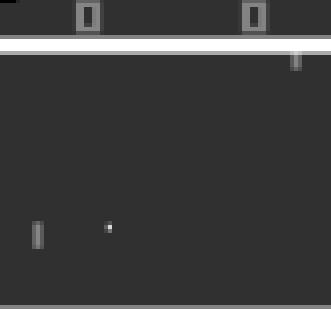}&
  \includegraphics[width=.09\textwidth]{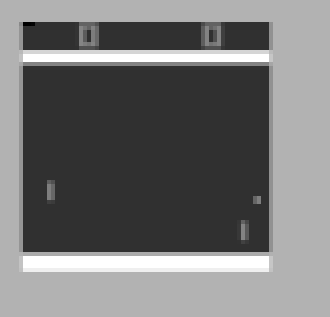}\\
  \includegraphics[width=.09\textwidth]{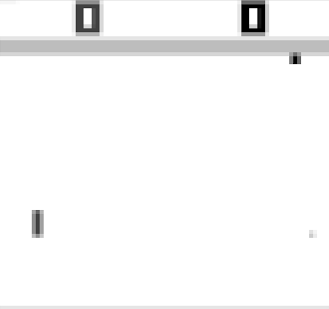}&
  \includegraphics[width=.09\textwidth]{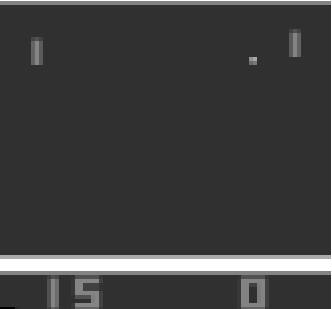}
\end{tabular}
}
\subfloat[Labyrinth games]{
\begin{tabular}{cc}
  \includegraphics[width=.15\textwidth]{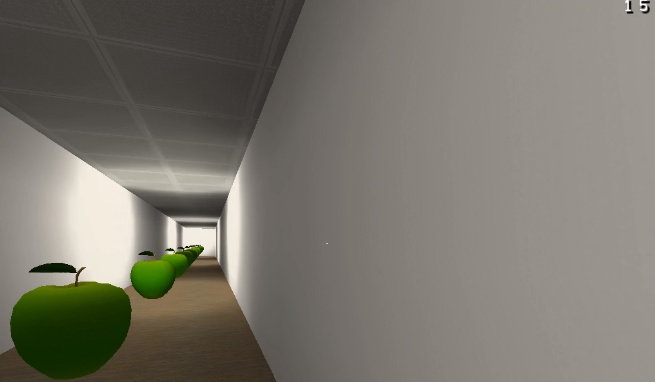}&
  \includegraphics[width=.15\textwidth]{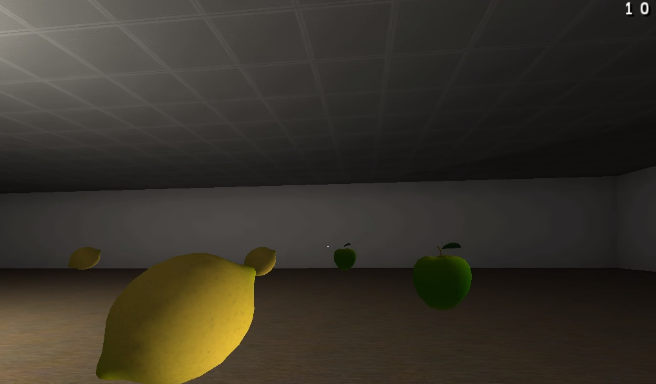}\\
  \includegraphics[width=.15\textwidth]{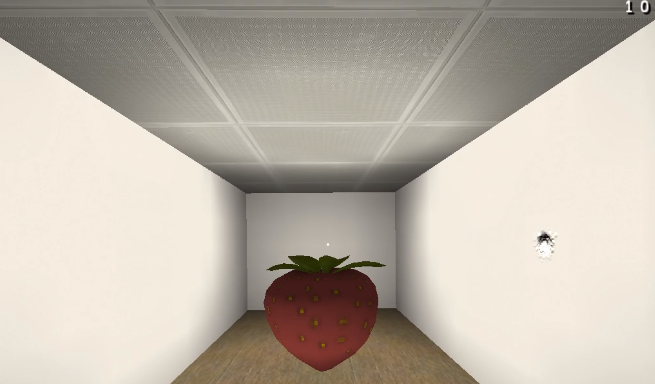}&
  \includegraphics[width=.15\textwidth]{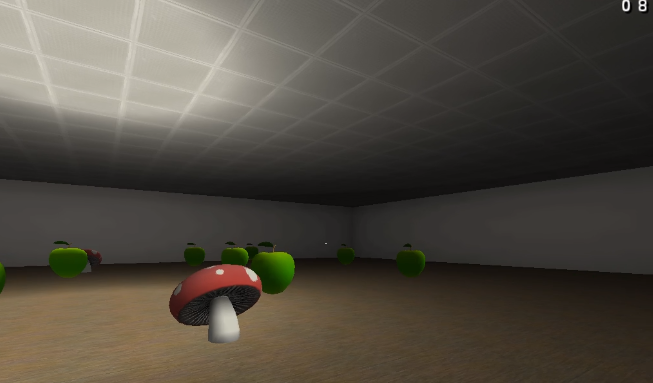}
\end{tabular}
}
\subfloat[Atari games]{
\begin{tabular}{cc}
  \includegraphics[width=.12\textwidth]{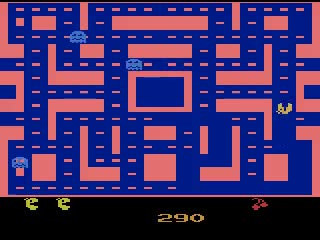}&
  \includegraphics[width=.12\textwidth]{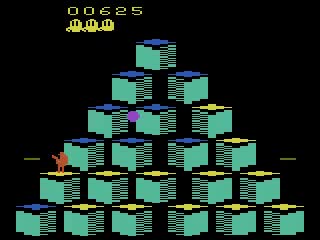}\\
  \includegraphics[width=.12\textwidth]{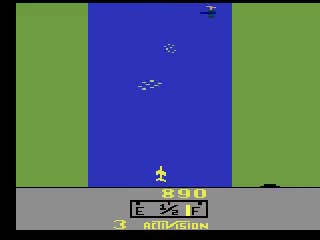}&
  \includegraphics[width=.12\textwidth]{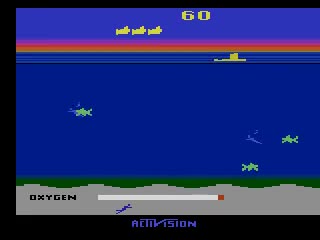}
\end{tabular}
}
\hfill
\caption{Samples from different task domains: (a) Pong variants include flipped, noisy, scaled, and recoloured transforms; (b) Labyrinth is a set of 3D maze games with diverse level maps and diverse positive and negative reward items; (c) Atari games offer a more challenging setting for transfer.}
\label{fig:datasets}
\end{figure}

We evaluate progressive networks across three different RL domains.
First, we consider synthetic versions of Pong,
altered to have visual or control-level
similarities. Next, we experiment broadly with random
sequences of Atari games and perform a feature-level transfer analysis.
Lastly, we demonstrate performance
on a set of 3D maze games. Fig.~\ref{fig:datasets} shows examples from selected tasks.

\subsection{Setup}

We rely on the Async Advantage Actor-Critic (A3C) framework introduced in
\citep{mnih2016a3c}.  Compared to DQN \citep{mnih-dqn-2015}, the model simultaneously learns
a policy and a value function
for predicting expected future rewards. A3C is trained on CPU using multiple threads
and has been shown to converge faster than DQN on GPU. This made it
a more natural fit for the
large amount of sequential experiments required for this work.

We report results by averaging the top 3 out of 25 jobs, each having
different seeds and random hyper-parameter sampling. Performance is evaluated
by measuring the area under the learning curve (average score per
episode during training), rather than final score. The \emph{transfer score}
is then defined as the relative performance of an architecture compared with
a single column baseline, trained only on the target task (baseline 1). We present transfer
score curves for selected source-target games, and summarize all such pairs in
\emph{transfer matrices}.
Models and baselines we consider are illustrated in Figure \ref{fig:baselines}.
Details of the experimental setup are provided in section 3 of the Appendix.

\begin{figure}[h]
  \centering
    \includegraphics[width=.85\textwidth]{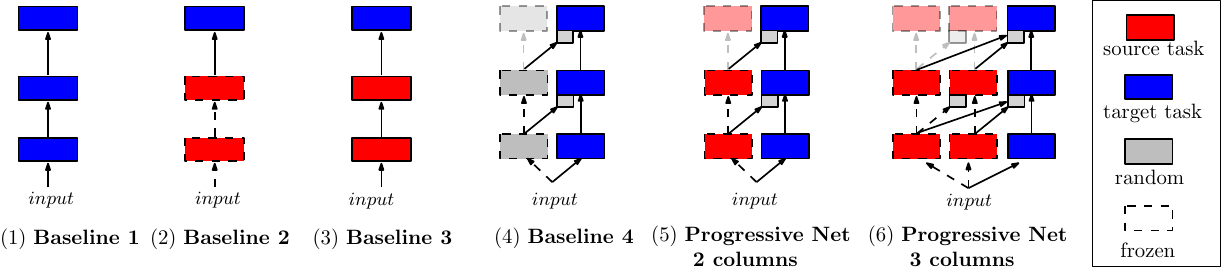}
    \caption{Illustration of different baselines and
      architectures. \emph{Baseline 1} is a single column trained on the target task;
\emph{baseline 2} is a single column, pretrained on a source task and
finetuned on the target task (output layer only); \emph{baseline 3} is
the same as baseline 2 but the whole model is finetuned; and
\emph{baseline 4} is a 2 column progressive architecture,
        with previous column(s) initialized randomly and frozen.
    }
    \label{fig:baselines}
\end{figure}

%% file: infinite_pong.tex
\subsection{Pong Soup}
\label{sec_pong}

\begin{figure}[h]
  \centering
    \includegraphics[width=1.\textwidth]{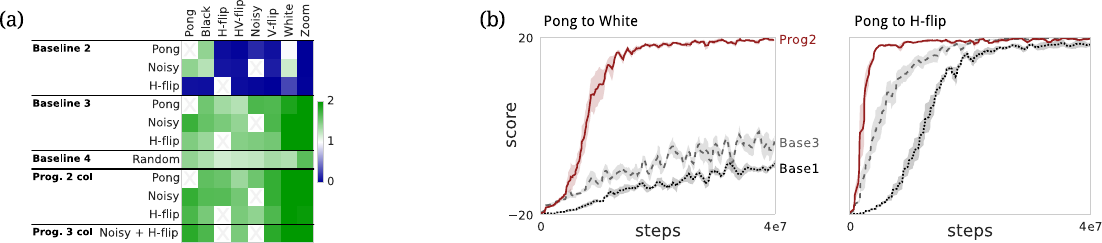}
    \caption{(a) Transfer matrix. Colours indicate transfer scores (clipped at 2).
      For progressive nets, the first column is trained on Pong, Noisy,
      or H-flip (table rows); the second column is trained on each of the
      other pong variants (table columns). (b) Example learning curves.}
    \label{fig:atari_results}
    \label{pong_results}
\end{figure}

The first evaluation domain is a set of synthetic variants of the
Atari game of Pong ("Pong Soup") where the visuals and gameplay have been
altered, thus providing a setting where we can be confident that there
are transferable aspects of the tasks.  The variants are
\emph{Noisy} (frozen Gaussian noise is added to the inputs);
\emph{Black} (black background); \emph{White} (white
background); \emph{Zoom} (input is scaled by 75\% and
translated); \emph{V-flip, H-flip, and VH-flip} (input is
horizontally and/or vertically flipped). Example frames are shown in
Fig. \ref{fig:datasets}.
The results of training two columns on the Pong variants, including
all relevant baselines are shown in Figure
\ref{pong_results}. Transfer scores are summarized over all target
tasks in Table~\ref{table:main}.

We can make
several observations from these results. Baseline 2 (single column, only output layer is finetuned; see Fig.~\ref{fig:baselines})
fails to learn the target task in most
experiments and thus has negative transfer. This approach is quite standard
in supervised learning settings, where features from
ImageNet-trained nets are routinely repurposed for new domains.
As expected, we observe high positive transfer with baseline 3 (single column, full finetuning),
a well established paradigm for transfer.
Progressive networks outperform this baseline however in terms of both median and mean score, with
the difference being more pronounced for the latter. As the mean is
more sensitive to outliers, this suggests that progressive networks are better
able to exploit transfer when transfer is possible (i.e. when source and target domains are compatible). Fig.~\ref{pong_results}~(b) lends
weight to this hypothesis, where progressive networks are shown to significantly
outperform the baselines for particular game pairs.
Progressive nets also compare
favourably to baseline 4,
confirming that progressive nets are indeed taking
advantage of the features learned in previous columns.

\textbf{Detailed analysis}

\begin{figure}[h]
  \centering
    \includegraphics[width=.95\textwidth]{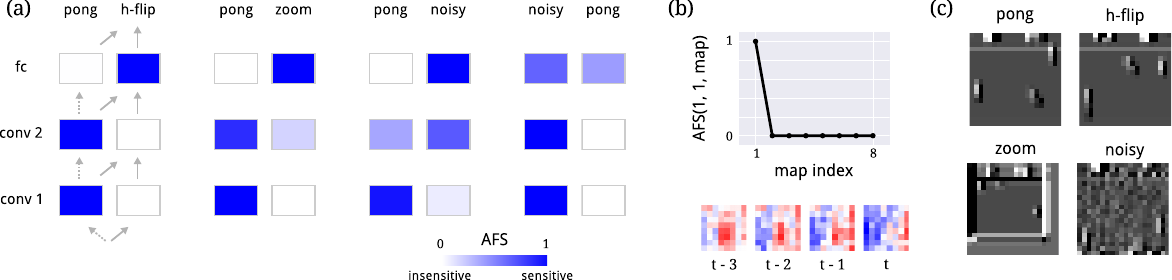}
    \caption{(a) Transfer analysis for 2-column nets on Pong variants.
      The relative sensitivity of the network's outputs on the columns
      within each layer (the AFS) is indicated by the darkness of shading.
      (b) AFS values for the 8 feature maps of
      conv. 1 of a 1-column Pong net. Only one feature map is
      effectively used by the net; the same map is also used by the
      2-column versions. Below: spatial filter components (red =
      positive, blue = negative). (c) Activation maps of the filter in (b) from example
      states of the four games.}
    \label{fig:pong_results_neil}
\end{figure}

We use the metric derived in Sec.~\ref{sec_transfer} to
analyse what features are being transferred between Pong variants.
We see that when switching from Pong to H-Flip, the network reuses the
same components of low and mid-level vision (the outputs of the two
convolutional layers; Figure \ref{fig:pong_results_neil}a). However, the fully connected layer must be
largely re-learned, as the policy relevant features of the task (the
relative locations/velocities of the paddle and ball) are now in a new
location. When switching from Pong to Zoom, on the other hand,
low-level vision is reused for the new task, but new mid-level vision
features are learned. Interestingly, only one low-level feature appears
to be reused:
(see Fig.~\ref{fig:pong_results_neil}b): this is a spatio-temporal
filter with a considerable temporal DC component. This appears
sufficient for detecting both ball motion and paddle position in the
original, flipped, and zoomed Pongs.

Finally, when switching from Pong to Noisy, some new low-level
vision is relearned. This is likely because the first layer
filter learned on the clean task is not sufficiently
tolerant to the added noise. In contrast, this problem does not apply
when moving from Noisy to Pong (Figure
\ref{fig:pong_results_neil}a, rightmost column), where all of vision
transfers to the new task.

%% file: two_columns.tex
\subsection{Atari Games}

We next investigate feature transfer between randomly
selected Atari games~\cite{bellemare13arcade}. This is an interesting question, because the visuals of
Atari games are quite different from each other, as are the controls
and required strategy. Though games like Pong and Breakout are
conceptually similar (both involve hitting a ball with a
paddle), Pong is vertically aligned while
Breakout is horizontal: a potentially insurmountable
feature-level difference. Other Atari game pairs have \emph{no} discernible
overlap, even at a conceptual level.

\begin{figure}[h]
  \centering
    \includegraphics[width=.95\textwidth]{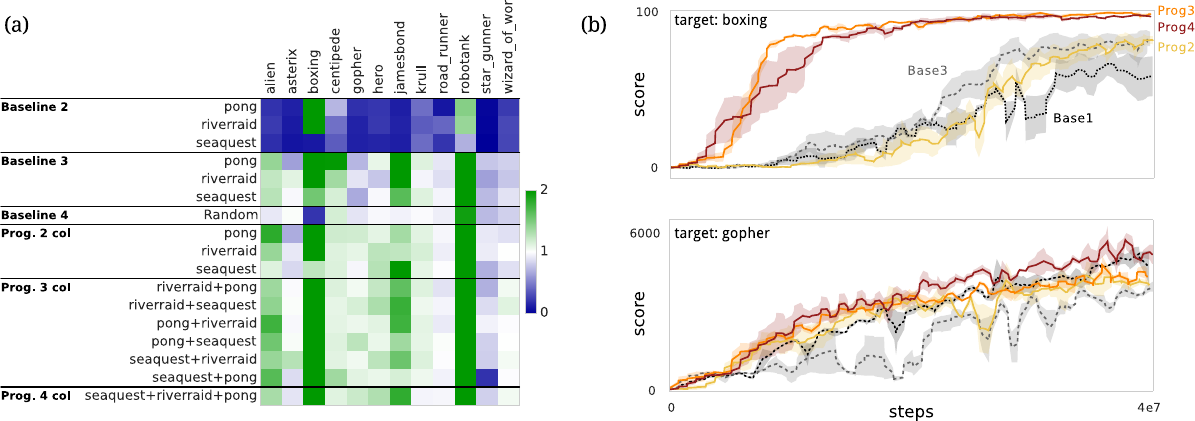}
    \caption{Transfer scores and example learning curves for Atari target games,
      as per Figure \ref{pong_results}.}
    \label{fig:atari_results}
\end{figure}

To this end we start by training single columns on three \emph{source}
games (Pong, River Raid, and Seaquest)
\footnote{Progressive columns having more than one ``source'' column are trained
sequentially on these source games, i.e.\ Seaquest-River Raid-Pong means column 1 is first trained
on Seaquest, column 2 is added afterwards and trained on River Raid, and then column 3 added and
trained on Pong.}
and assess if the learned features transfer to a different subset of
randomly selected \emph{target} games (Alien, Asterix, Boxing, Centipede,
Gopher, Hero, James Bond, Krull, Robotank, Road Runner, Star Gunner,
and Wizard of Wor). We evaluate progressive networks with 2, 3 and 4
columns, comparing to the baselines of Figure \ref{fig:baselines}).
The transfer matrix and selected transfer curves are shown in
Figure~\ref{fig:atari_results}, and the results summarized in
Table~\ref{table:main}.

Across all games, we observe from Fig.~\ref{fig:atari_results},
that progressive nets result in
positive transfer in 8 out of 12 target tasks, with only two cases
of negative transfer. This compares favourably to baseline 3, which yields
positive transfer in only 5 of 12 games. This trend is reflected in
Table~\ref{table:main}, where progressive networks convincingly outperform
baseline 3 when using additional columns. This is especially promising as
we show in the Appendix that progressive network use a diminishing amount of
capacity with each added column, pointing a clear path to online compression
or pruning as a means to mitigate the growth in model size.

Now consider the specific sequence
\textit{Seaquest}-to-\textit{Gopher}, an example of two dissimilar games. Here, the
pretrain/finetune paradigm (baseline 3) exhibits negative transfer, unlike
progressive networks (see Fig.\ref{fig:atari_results}b, bottom), perhaps
because they are more able to ignore the irrelevant features. For the sequence
\textit{Seaquest[+River Raid][+Pong]}-to-\textit{Boxing}, using additional
columns in the progressive networks can yield a significant increase in
transfer (see Fig.~\ref{fig:atari_results}b, top).

\begin{table}[t]
\begin{center}
\small
\begin{tabular}{@{}lrrrrrr@{}}
 \toprule
& \multicolumn{2}{c}{\textbf{Pong Soup}}
& \multicolumn{2}{c}{\textbf{Atari}}
& \multicolumn{2}{c}{\textbf{Labyrinth}}\\
                    & Mean (\%)    & Median (\%)  & Mean (\%)    & Median (\%)  & Mean (\%)     & Median (\%)  \\
 \midrule
  Baseline 1        & 100          & 100          & 100          & 100          & 100           & 100          \\
  Baseline 2        & 35           & 7            & 41           & 21           & 88            & 85           \\
  Baseline 3        & 181          & 160          & 133          & 110          & 235           & 112          \\
  Baseline 4        & 134          & 131          & 96           & 95           & 185           & 108          \\
  Progressive 2 col & 209          & 169          & 132          & 112          & \textbf{491}  & \textbf{115} \\
  Progressive 3 col & \textbf{222} & \textbf{183} & 140          & 111          & ---           & ---          \\
  Progressive 4 col & ---          & ---          & \textbf{141} & \textbf{116} & ---           & ---          \\

 \bottomrule
\end{tabular}
\end{center}
\caption{Transfer percentages in three domains. Baselines are defined in
    Fig.~\ref{fig:baselines}.}
\label{table:main}
\end{table}

%% file: atari_analysis.tex
\textbf{Detailed Analysis}

\begin{figure}[h]
  \centering \includegraphics[width=.95\textwidth]{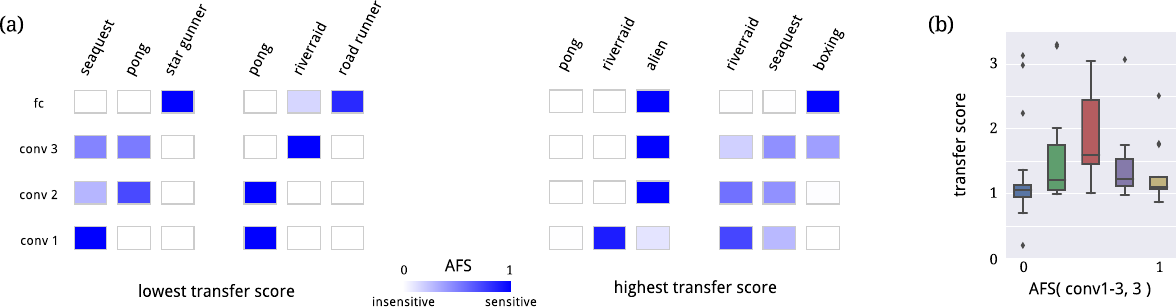} \caption{(a)
    AFS scores for 3-column nets with lowest (left) and highest
    (right) transfer scores on the 12 target Atari games. (b) Transfer
    statistics across 72 three-column nets, as a function of the
    mean AFS across the three convolutional layers of the new
    column (i.e.\ how much new vision is learned). } \label{fig:atari3_results_neil}
\end{figure}

Figure \ref{fig:atari_results} demonstrates that both positive and
negative transfer is possible with progressive nets. To differentiate
these cases, we consider the Average Fisher Sensitivity for the 3 column
case (e.g., see Fig.~\ref{fig:atari3_results_neil}a). A clear pattern
emerges amongst these and other examples: the most negative transfer
coincides with complete dependence on the convolutional layers of the
previous columns, and no learning of new visual features in the new
column. In contrast, the most positive transfer occurs when the
features of the first two columns are \textit{augmented} by
new features. The statistics across all 3-column nets
(Figure \ref{fig:atari3_results_neil}b) show that positive transfer in
Atari occurs at a "sweet spot" between heavy reliance on features from
the source task, and heavy reliance on all new features for the
target task.

At first glance, this result appears unintuitive: if a progressive net
finds a valuable feature set from a source task,
shouldn't we expect a high degree of transfer?
We offer two hypotheses. First, this may
simply reflect an optimization difficulty, where the source features offer
fast convergence to a poor local minimum. This is a known
challenge in transfer learning \cite{AAAIMag11-Taylor}: learned source
tasks confer an inductive bias that can either help or hinder in different cases.
Second, this may reflect a problem of
exploration, where the transfered representation is "good enough" for
a functional, but sub-optimal policy.

%% file: labyrinth.tex
\subsection{Labyrinth}

The final experimental setting for progressive networks is Labyrinth,
a 3D maze environment where the inputs are rendered images granting partial
observability and the agent outputs discrete actions,
including looking up, down, left, or right and moving forward,
backwards, left, or right. The tasks as well as the level maps are
diverse and involve getting positive scores for `eating' good items
(apples, strawberries) and negative scores for eating bad items
(mushrooms, lemons). Details can be found in the appendix. While there is conceptual and
visual overlap between the different tasks, the tasks present a
challenging set of diverse game elements (Figure
\ref{fig:datasets}).

\begin{figure}[h]
  \centering
    \includegraphics[width=\textwidth]{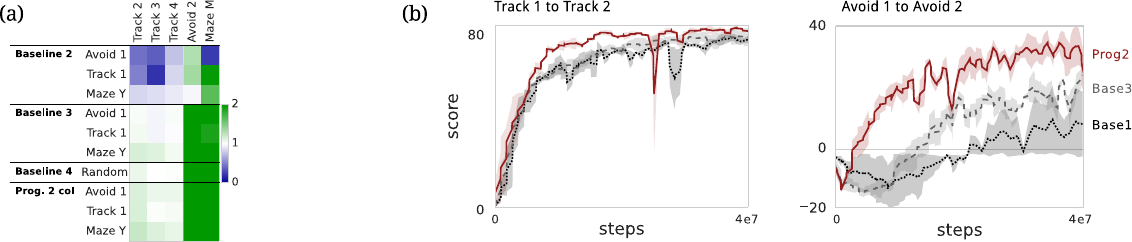}
    \caption{Transfer scores and example learning curves for Labyrinth tasks. Colours indicate transfer (clipped at 2). The learning curves show two examples of two-column progressive performance vs. baselines 1 and 3.}
    \label{fig:lab}
\end{figure}

As in the other domains, the progressive approach yields more positive transfer than any of the baselines (see Fig.~\ref{fig:lab}a and Table~\ref{table:main}). We observe less transfer on the Seek Track levels, which have dense reward items throughout the maze and are easily learned. Note that even for these easy cases, baseline 2 shows negative transfer because it cannot learn new low-level visual features, which are important because the reward items change from task to task. The learning curves in Fig.~\ref{fig:lab}b exemplify the typical results seen in this domain: on simpler games, such as Track 1 and 2, learning is rapid and stable by all agents. On more difficult games, with more complex game structure, the baselines struggle and progressive nets have an advantage.

%% file: discussion.tex
\section{Conclusion}

Continual learning, the ability to accumulate and transfer knowledge to new domains, is a core characteristic of intelligent beings. \textit{Progressive neural networks} are a stepping stone towards continual learning, and this work has demonstrated their potential through experiments and analysis across three RL domains, including Atari, which contains orthogonal or even adversarial tasks. We believe that we are the first to show positive transfer in deep RL agents within a continual learning framework. Moreover, we have shown that the progressive approach is able to effectively exploit transfer for compatible source and task domains; that the approach is robust to harmful features learned in incompatible tasks; and that positive transfer increases with the number of columns, thus corroborating the constructive, rather than destructive, nature of the progressive architecture.

%% file: app_perturbation.tex
\section{Perturbation Analysis}

We explored two related methods for analysing transfer in progressive networks.
One based on Fisher information yields the Average Fisher Sensitivity (AFS)
and is described in Section 3 of the paper. We describe
the second method based on perturbation analysis in this appendix, as it proved
too slow to use at scale. Given its intuitive appeal however, we provide
details of the method along with results on Pong Variants (see
Section 5.2), as a means to corroborate the AFS score.

Our perturbation analysis aims to estimate which components of the source
columns materially contribute to the performance of the final column on the
target tasks. To this end, we injected Gaussian noise into each of the (post-ReLU) hidden
representations, with a new sample on every forward pass, and calculated the
average effect of these perturbations on the game score over 10 episodes.  We
did this at a coarse scale, by adding noise across all features of a given
layer, though a fine scale analysis is also possible per feature (map).
In order to be invariant to any arbitrary scale factors in the network weights,
we scale the noise variance proportional to the variance of the activations in
each feature map and fully-connected neuron. Scaling the variance in this
manner is analogous to computing the Fisher w.r.t. normalized activations for
the AFS score.

\begin{figure}[h]
  \centering
    \includegraphics[width=.95\textwidth]{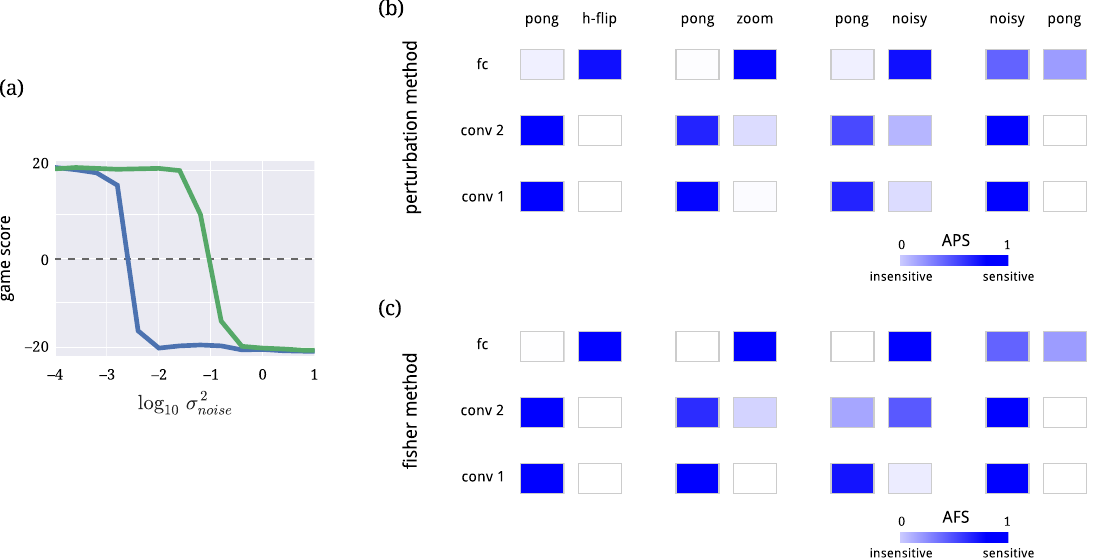}
    \caption{(a) Perturbation analysis for the two second-layer
      convolutional representations in the two columns of the
      Pong/Pong-noise net. Blue: adding noise to second convolutional layer from column 1;
      green: from column 2. Grey line determines critical noise
      magnitude for each representation, $\sigma_i^2$.
      (b-c) Comparison of per-layer sensitivities obtained using the APS
      method (b) and the AFS method (c; as per main text).
      These are highly similar.}
    \label{fig:app_afs_vs_aps}
\end{figure}

Define ${\Lambda_i^{(k)}}=1/\sigma_i^{2(k)}$ as the precision of the noise
injected at layer $i$ of column $k$, which results in a $50\%$ drop in
performance. The Average Perturbation Sensitivity (APS) for this layer is
simply:
\begin{align}
    \text{APS}(i,k) = \frac{\Lambda_i^{(k)}}{\sum_k \Lambda_i^{(k)}}
\end{align}
Note that this value is normalized across columns for a given layer. The APS
score can thus be interpreted as the responsibility of each column in a given
layer to final performance.
The APS score of 2-column progressive networks trained on Pong Variants is
shown in Fig\ref{fig:app_afs_vs_aps} (b). These clearly corroborate the
AFS shown in (c).

%% file: app_compression.tex
\section{Compressibility of Progressive Networks}

As described in the main text, one of the limitations of progressive networks is the growth in the size of the network with added tasks. In the basic approach we pursue in the main text, the number of hidden units and feature maps grows linearly with the number of columns, and the number of parameters grows quadratically.

\begin{figure}[h]
  \centering
    \includegraphics[width=.95\textwidth]{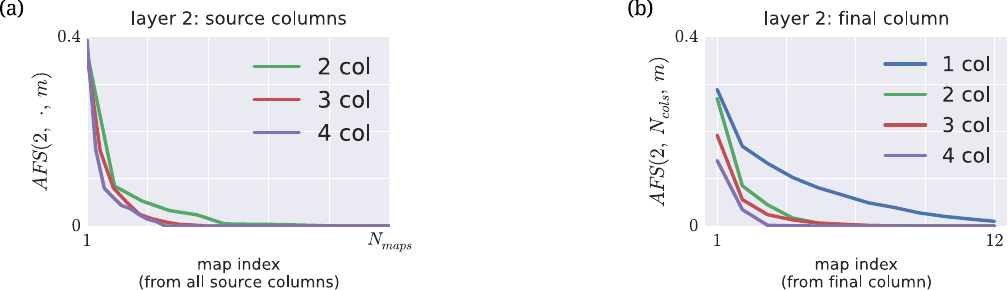}
    \caption{(a) Spectra of AFS values (for layer 2) across all feature maps from source columns, for the Atari dataset. The spectra show the range of AFS values, and are averaged across networks. While the 2 column / 3 column / 4 column nets all have different values of $N_{maps}$ (here, 12, 24, and 36 respectively), these have been dilated to fit the same axis to show the proportional use of these maps. (b) Spectra of AFS values (for layer 2) for the feature maps from only the final column. 
      }
    \label{fig:app_compression}
\end{figure}

Here, we sought to determine the degree to which this full capacity is actually used by the network. We leveraged the Average Fisher Sensitivity measure to study how increasing the number of columns in the Atari task set changes the need for additional resources. In Figure \ref{fig:app_compression}a, we measure the average fractional use of \textit{existing} feature maps in a given layer (here, layer 2). We do this for each network by concatenating the per-feature-map AFS values from all source columns in this layer, sorting the values to produce a spectrum, and then averaging across networks. We find that as the number of columns increases, the average spectrum becomes sparser: the network relies on a smaller proportion of features from the source columns. Similar results were found for all layers. 

Similarly, in Figure \ref{fig:app_compression}b, we measure the capacity required in the final added column as a function of the total number of columns. Again, we measure the spectrum of AFS values in an example layer, but here from only the final column. As the progressive network grows, the new column's features are both less important overall (indicated by the declining area under the graph), and have a sparser AFS spectrum. Combined, these results suggest that significant pruning of lateral connections is possible, and the quadratic growth of parameters might be contained.

%% file: app_setup.tex
\section{Setup Details}
\label{sec:appendix_jobs_detail}

In our grid we sample hyper-parameters from categorical distributions:
\begin{itemize}
  \item Learning rate was sampled from $\{10^{-3}, 5\cdot 10^{-4}, 10^{-4}\}$.
  \item Strength of the entropy regularization from $\{10^{-2}, 10^{-3}, 10^{-4}\}$
  \item Gradient clipping cut-off from $\{20, 40\}$
  \item scalar multiplier on the lateral feature is initialized randomly to one from $\{1, 10^{-1}, 10^{-2}\}$
\end{itemize}

For the Atari experiments we used a model with 3 convolutional layers followed by a fully connected
layer and from which we predict the policy and value function. The convolutional layers are as
follows. All have 12 feature maps. The first convolutional layer has a kernel of size 8x8 and a stride
of 4x4. The second layer has a kernel of size 4 and a stride of 2. The last convolutional layer has
size 3x4 with a stride of 1.  The fully connected layer has 256 hidden units.

Learning follows closely the paradigm described in \citep{mnih2016a3c}. We use 16 workers and the
same RMSProp algorithm without momentum or centring of the variance. The score for each point of
a training curve is the average over all the episodes the model gets to finish in $25e4$ environment steps.

The whole experiments are run for a maximum of $1.6e8$ environment step. The agent has an action repeat of 4 as
in \cite{mnih2016a3c}, which means that for 4 consecutive steps the agent will use the same action picked at the
beginning of the series. For this reason through out the paper we actually report results in terms of agent
perceived steps rather than environment steps. That is, the maximal number of agent perceived step that we do
for any particular run is $4e7$.

%% file: app_curves.tex
\section{Learning curves}
\label{sec:appendix_curves}

\begin{figure}
     \begin{tabular}{ccc}
        \includegraphics[width=.33\textwidth]{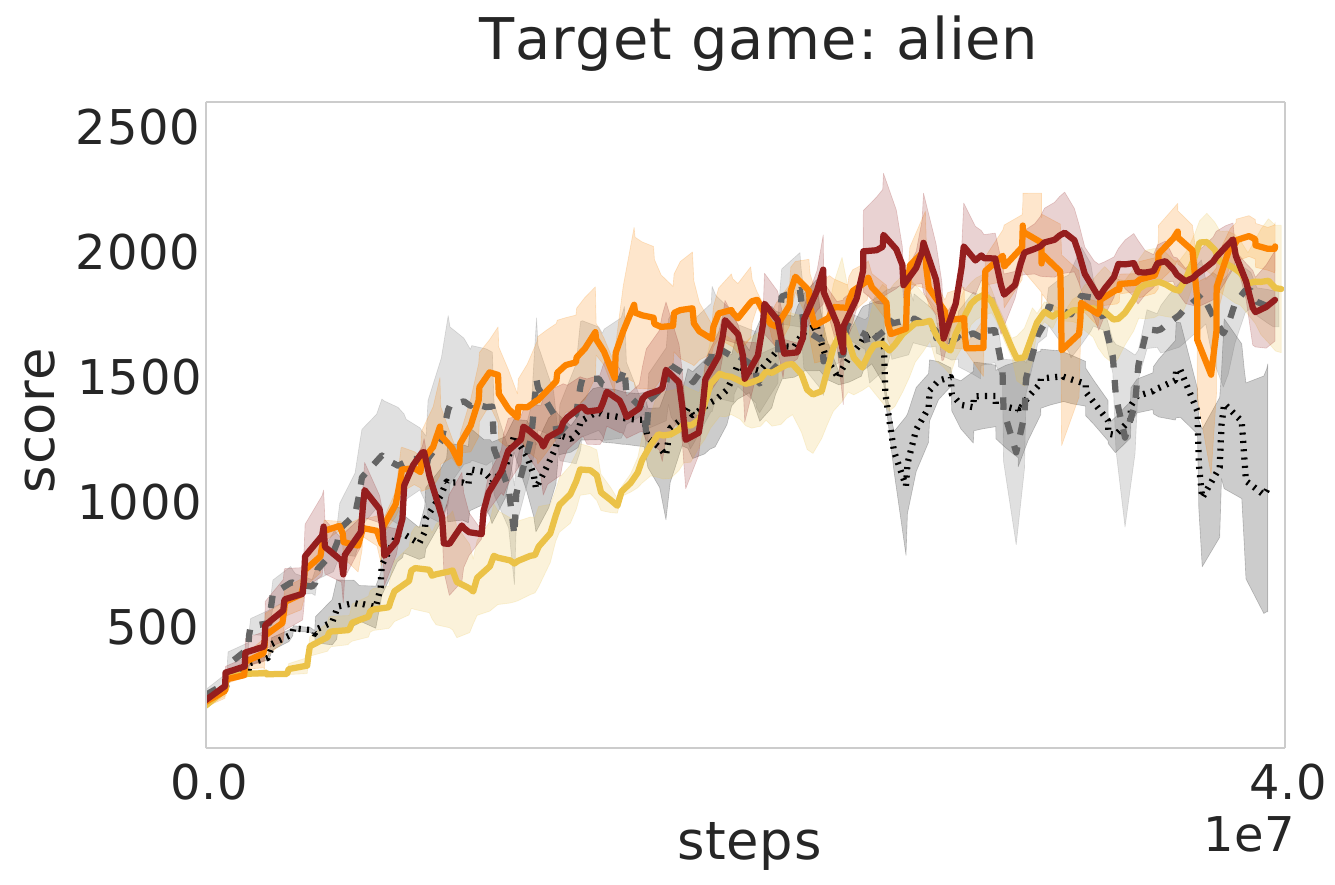} &
        \includegraphics[width=.33\textwidth]{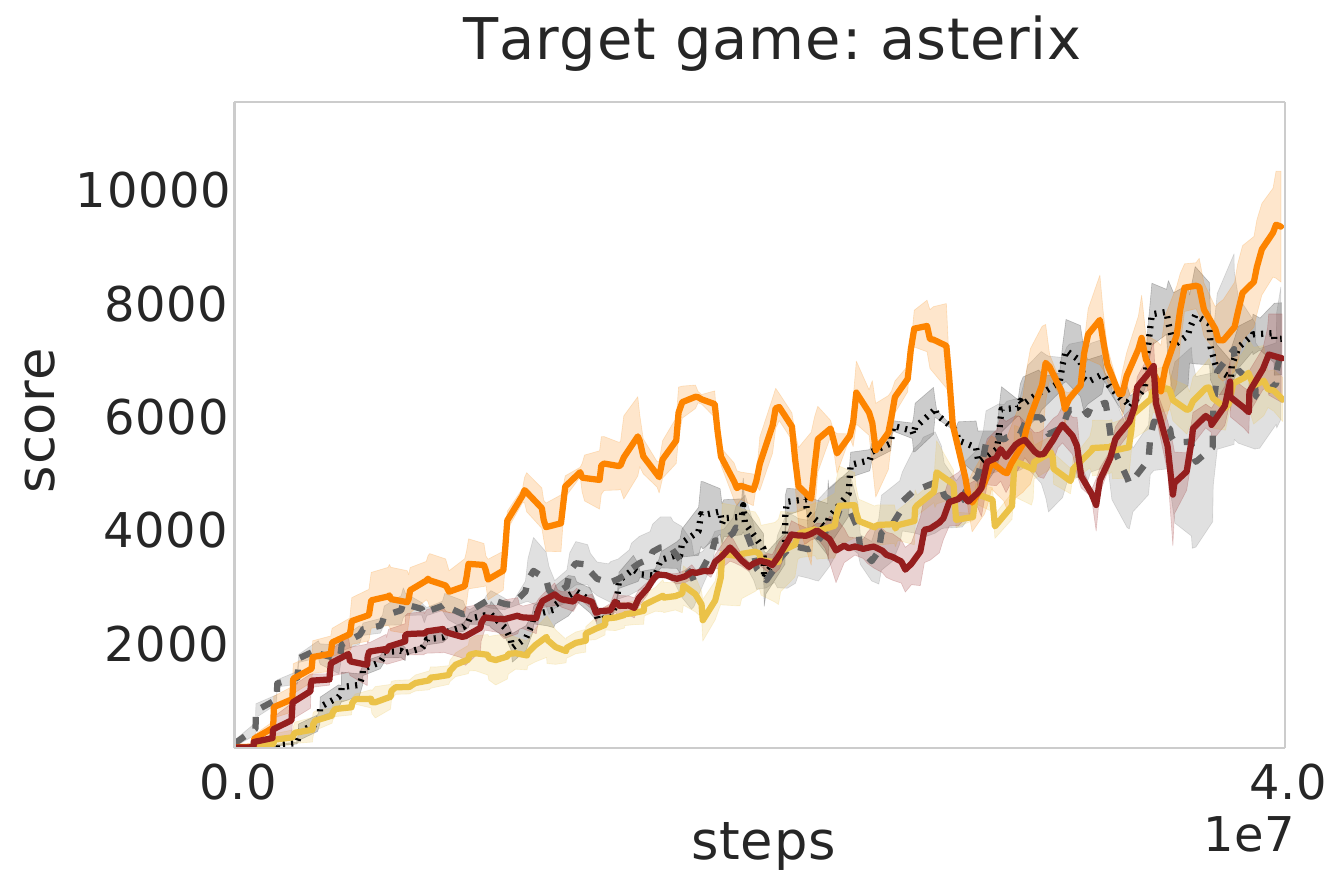} &
        \includegraphics[width=.33\textwidth]{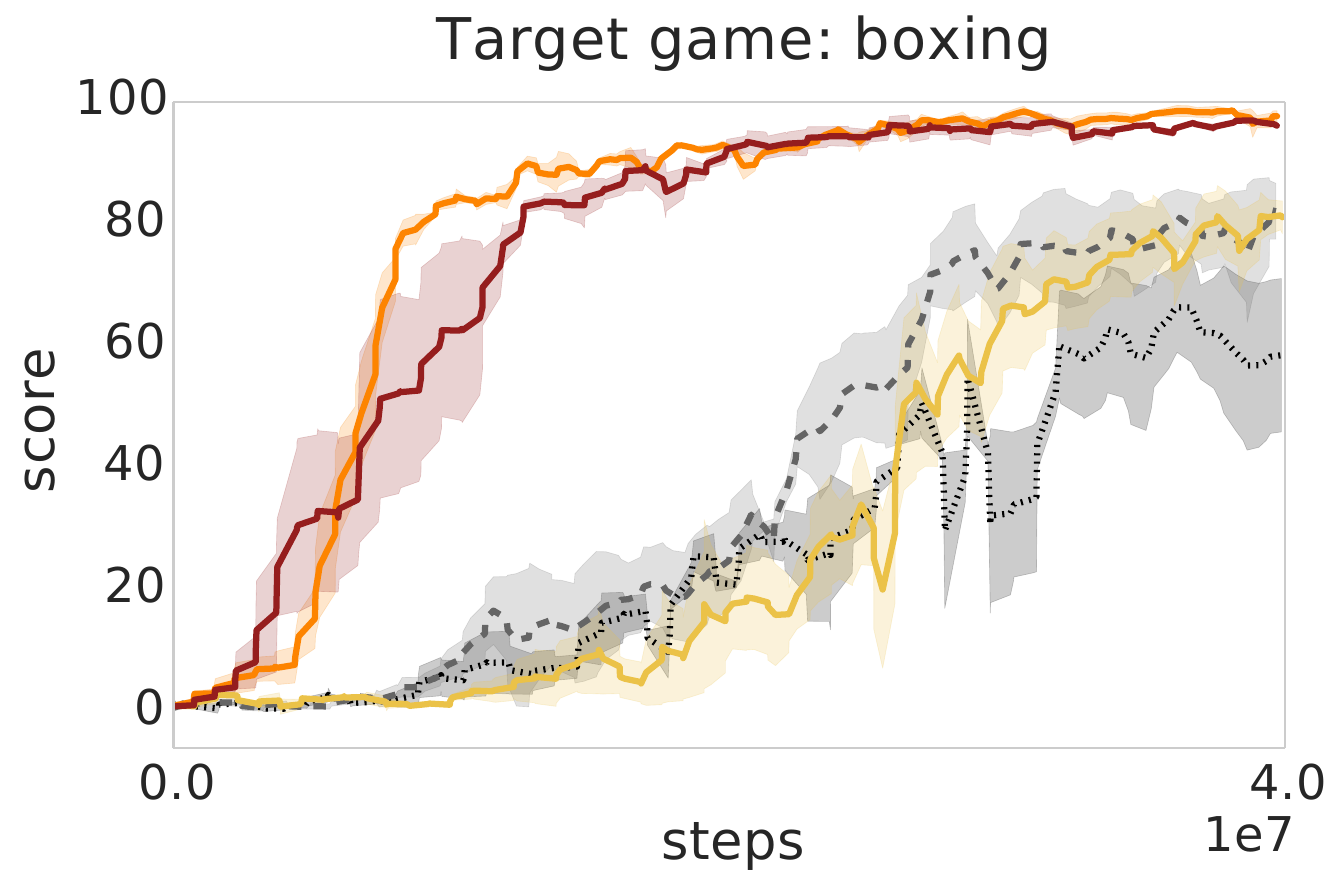} \\

        \includegraphics[width=.33\textwidth]{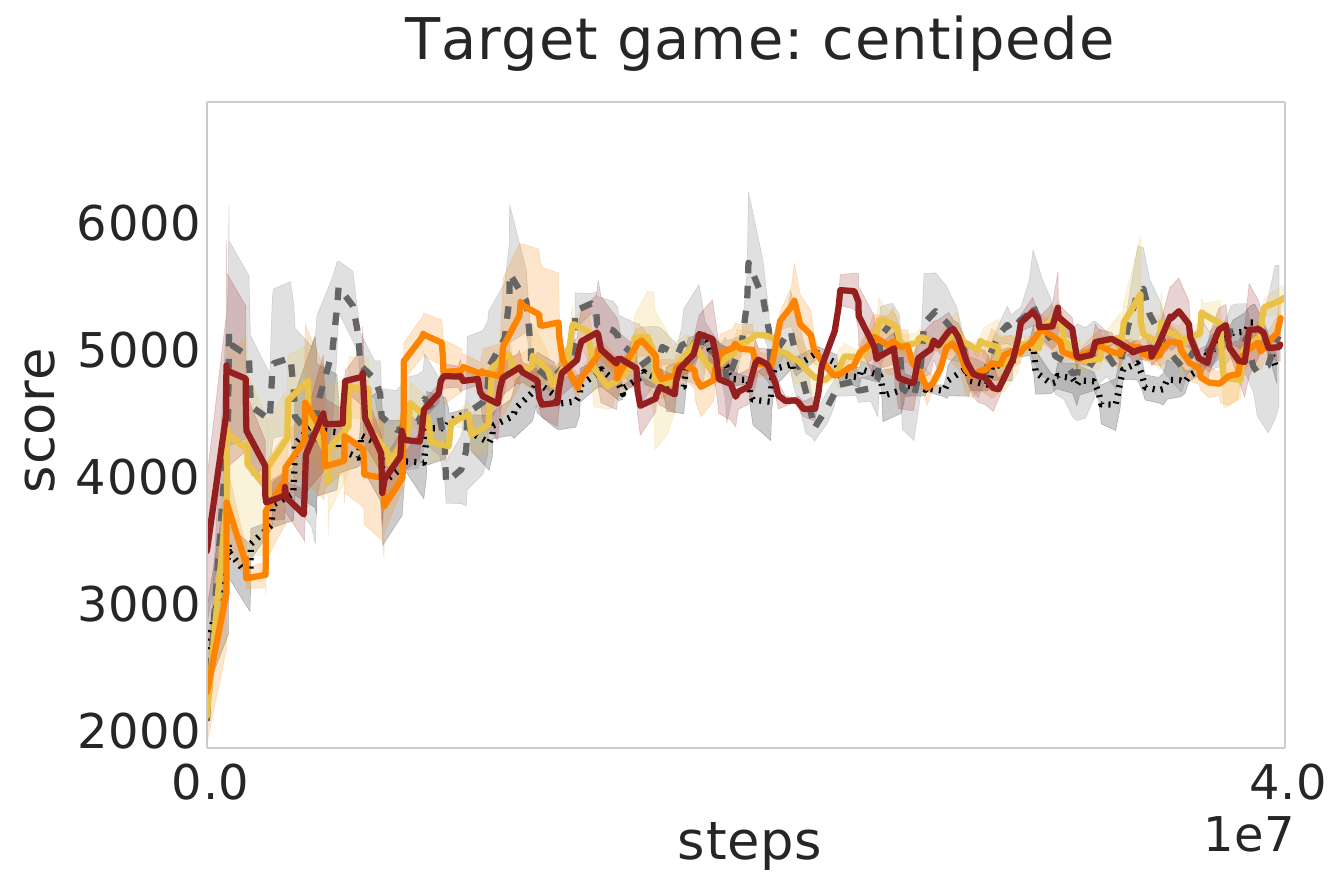} &
        \includegraphics[width=.33\textwidth]{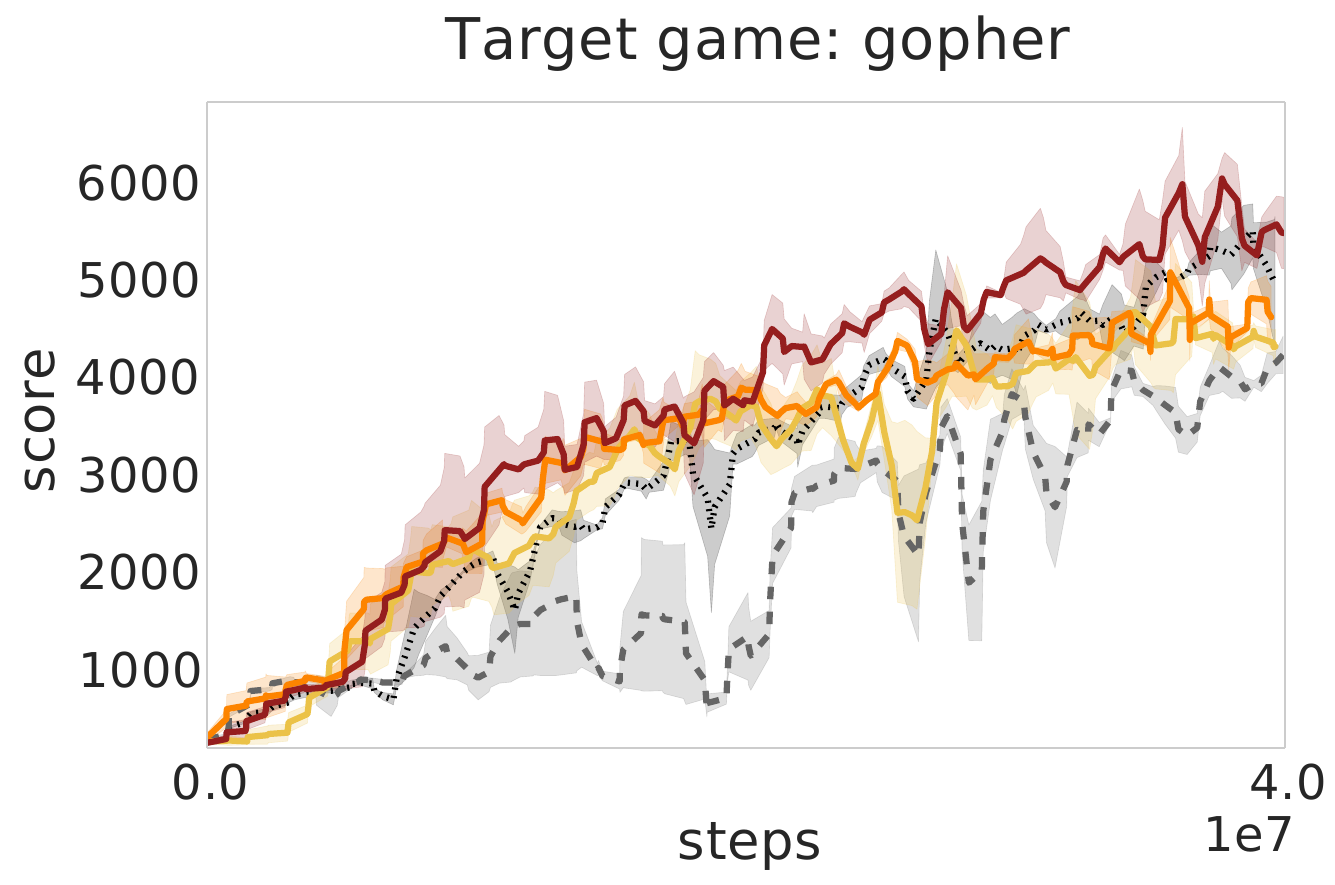} &
        \includegraphics[width=.33\textwidth]{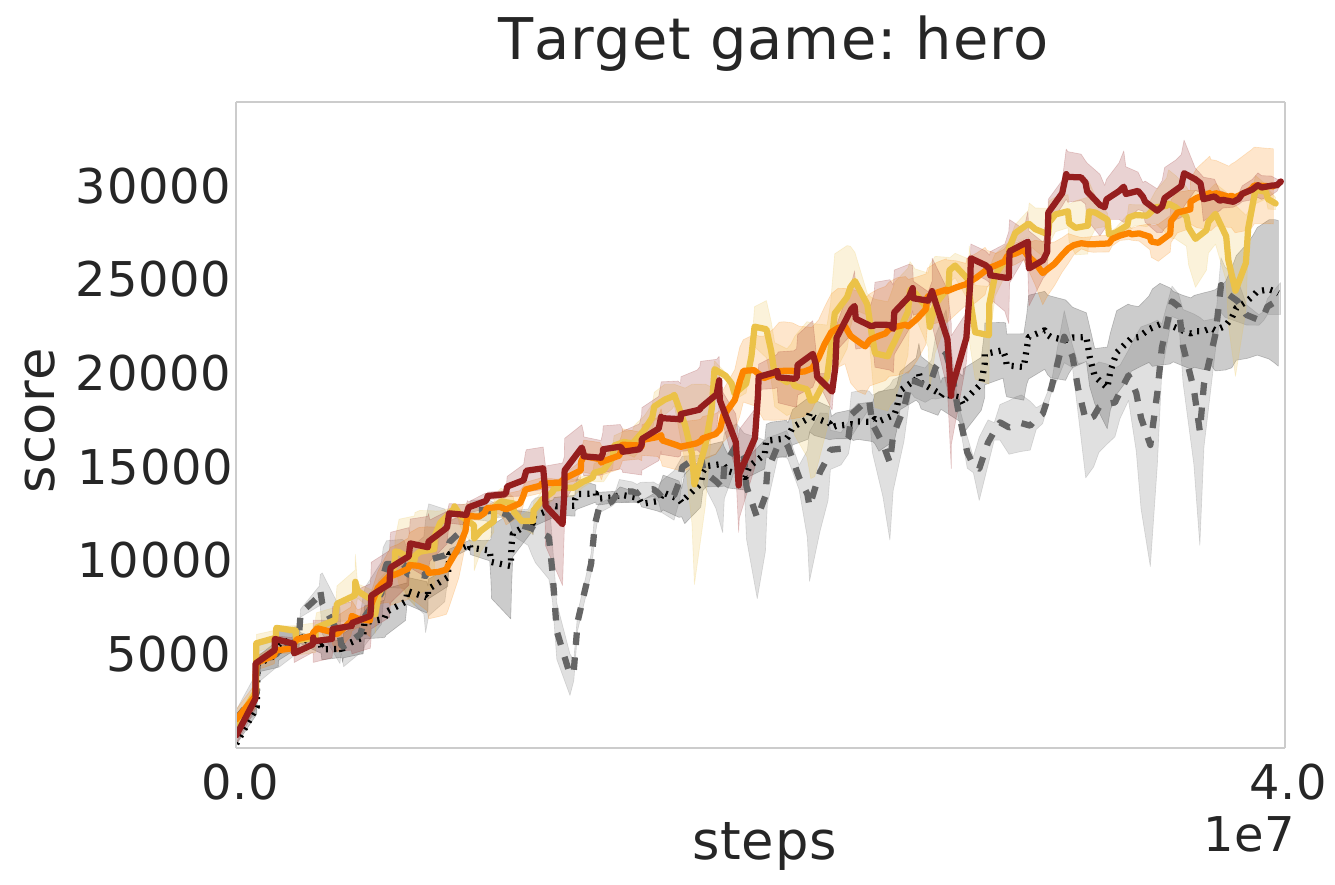} \\

        \includegraphics[width=.33\textwidth]{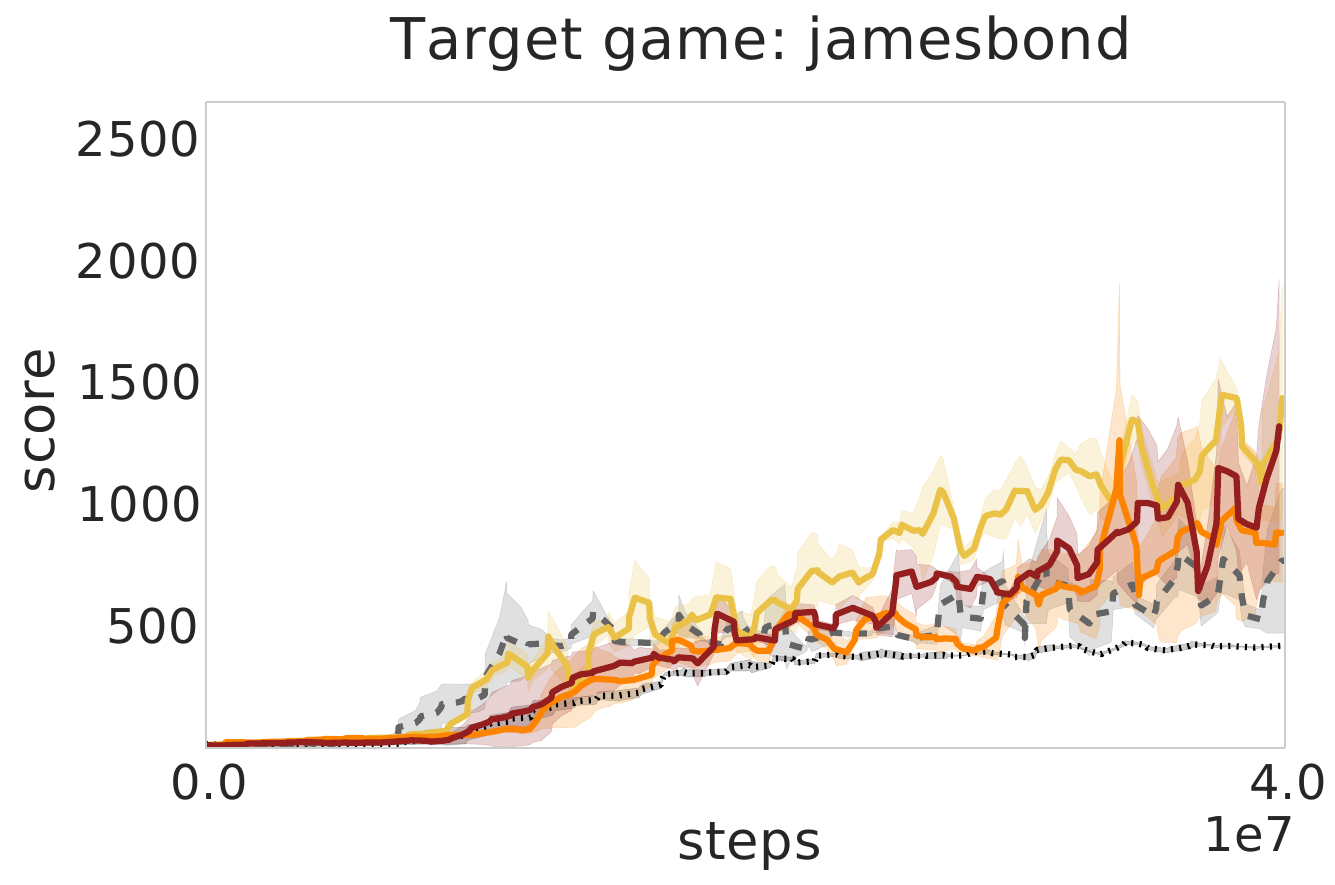} &
        \includegraphics[width=.33\textwidth]{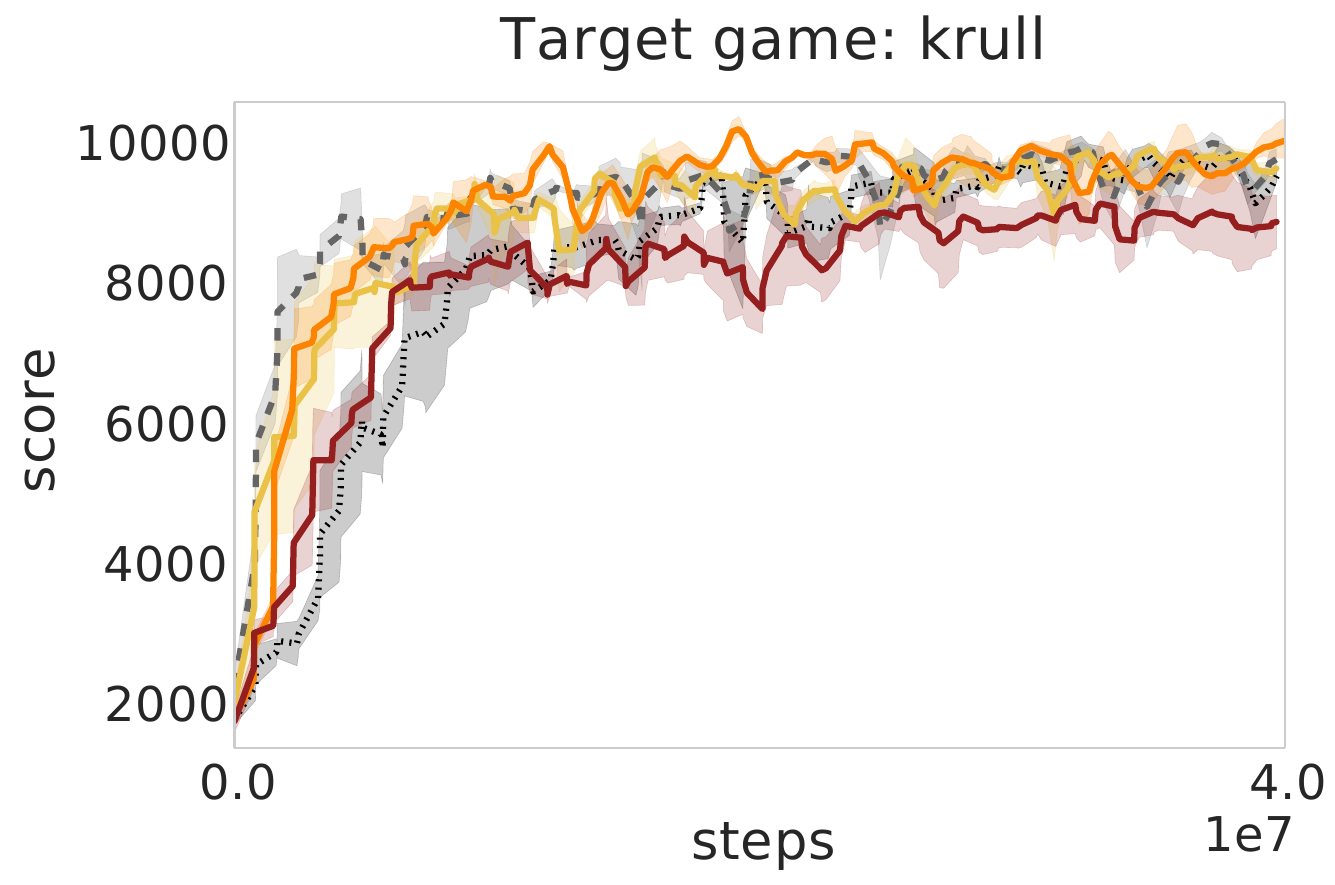} &
        \includegraphics[width=.33\textwidth]{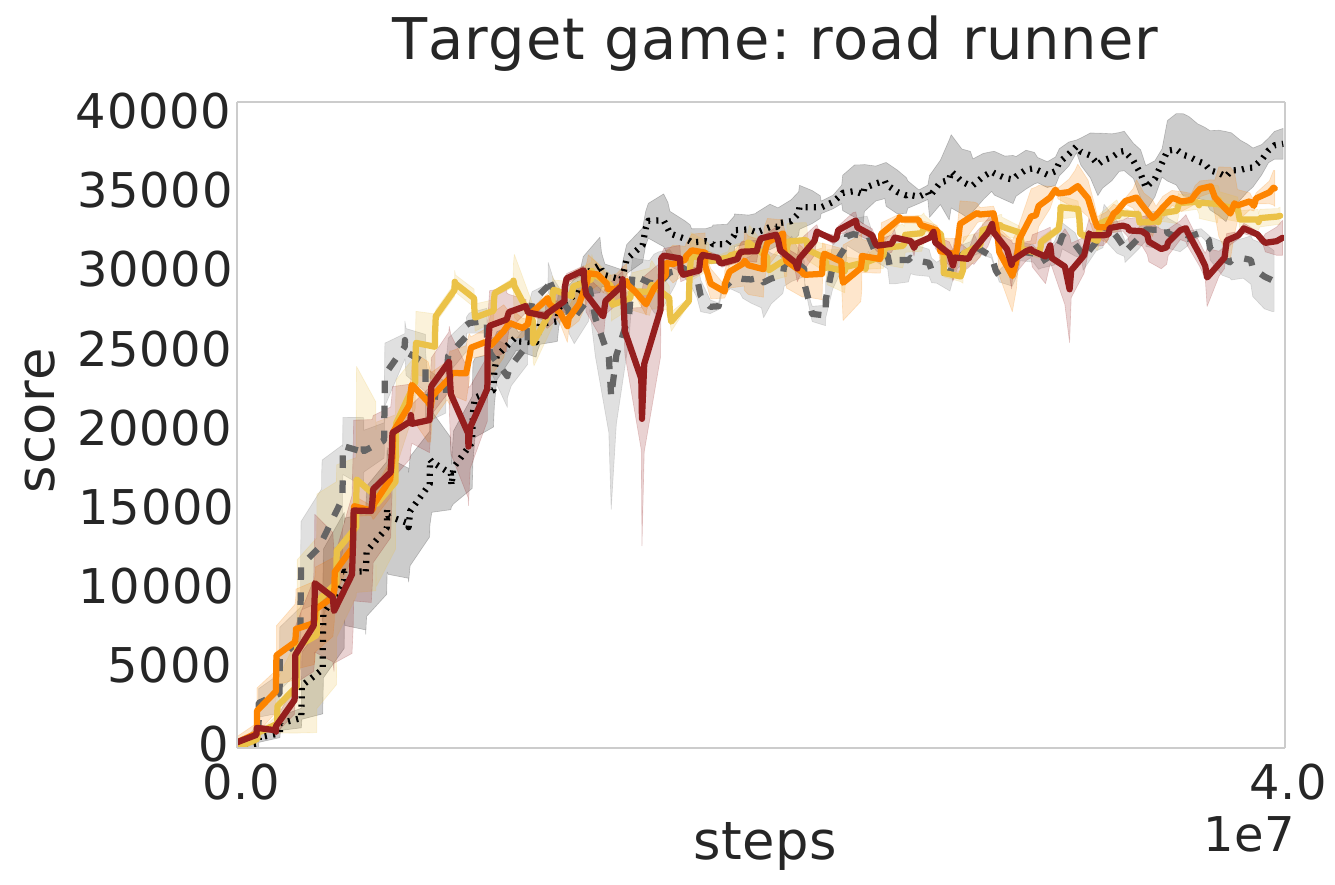} \\

        \includegraphics[width=.33\textwidth]{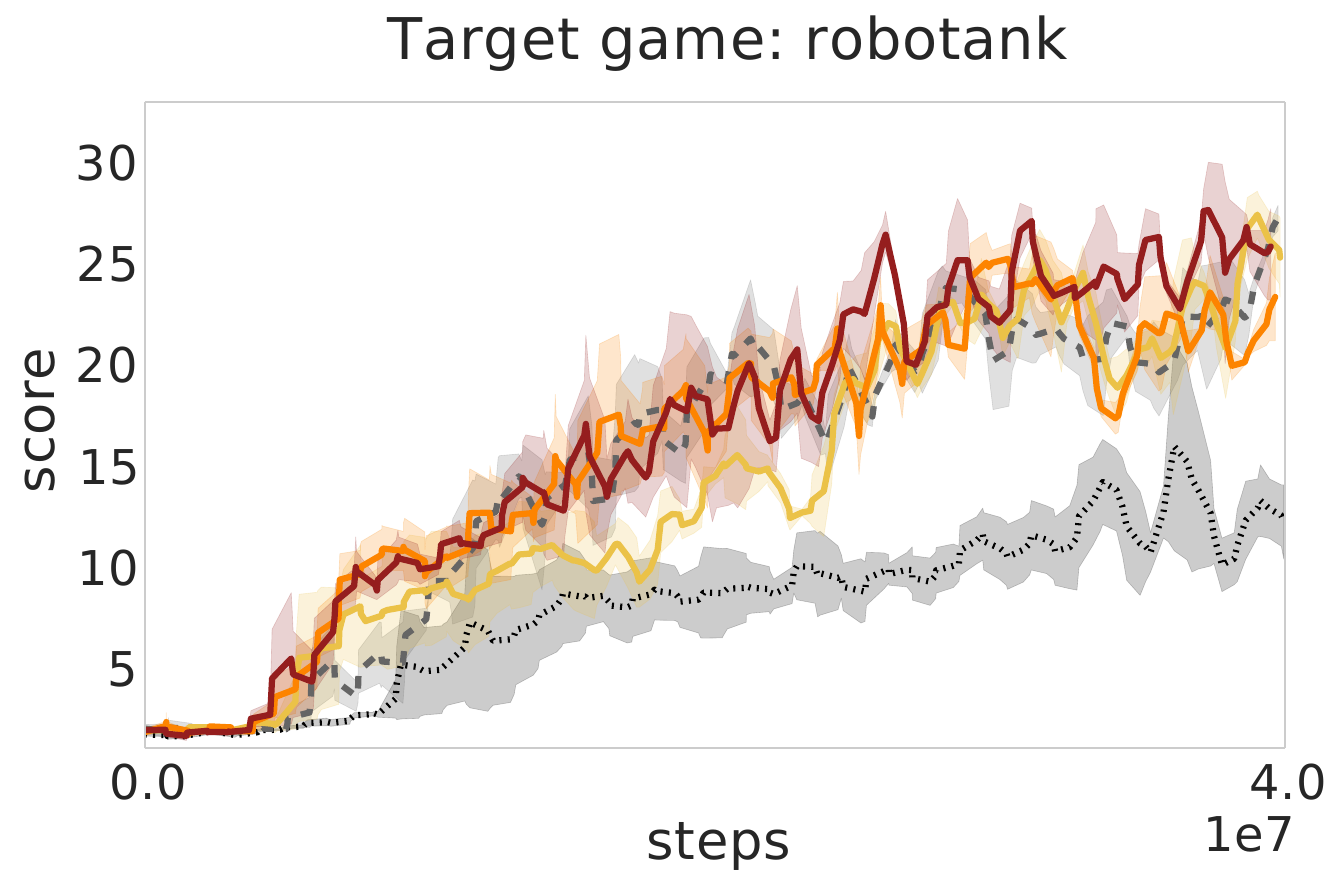} &
        \includegraphics[width=.33\textwidth]{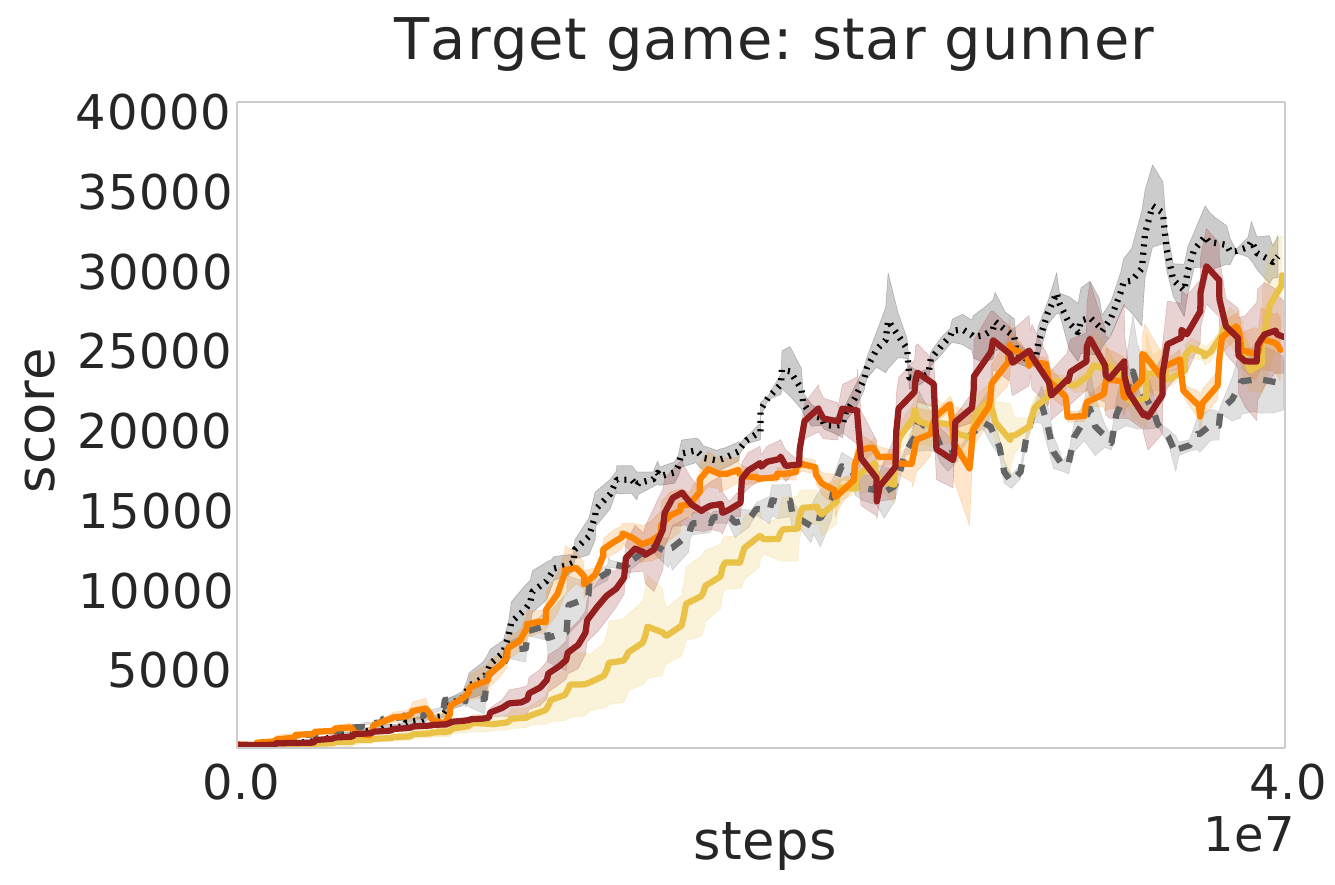} &
        \includegraphics[width=.33\textwidth]{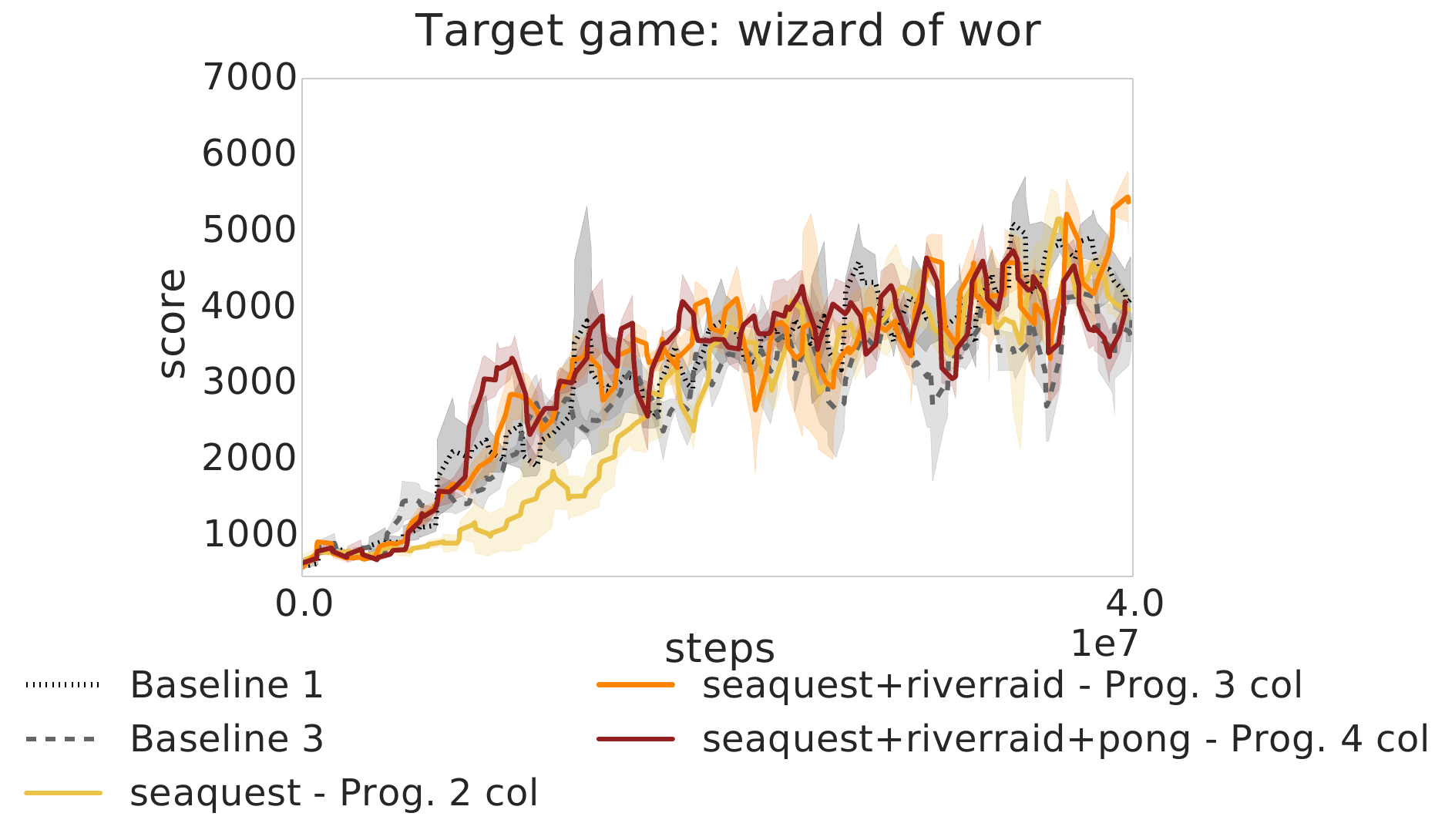} \\
    \end{tabular}
    \caption{Training curves for transferring to the \textit{target} games after seeing first Seaquest followed by River Raid and lastly Pong. For the baselines,
    the \textit{source} game used for pretraining is Seaquest.}
    \label{fig:app_plot}
\end{figure}

Figure \ref{fig:app_plot} shows training curves for all the \textit{target} games in the Atari domain.
We plot learning curves for two column, three column and four column progressive networks alongside Baseline 3 (gray dashed line), a model pretrained on Seaquest and then finetuned on the particular \textit{target} game and Baseline 1 (gray dotted line), where a single column is trained on the \textit{source} game Seaquest.

We can see that overall baseline 3 performs well. However there are situations when having features learned from more previous task actually helps with transfer (e.g. when \textit{target} game is Boxing).

\begin{figure}
     \begin{tabular}{cccccccc}
	Target: Pong & Target: Black \\
        \includegraphics[width=.44\textwidth]{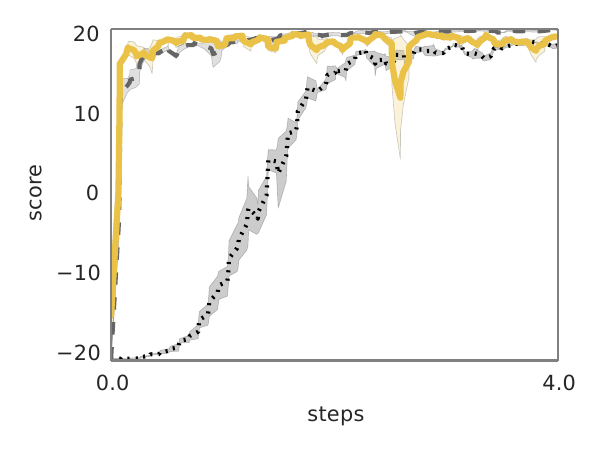} &
        \includegraphics[width=.44\textwidth]{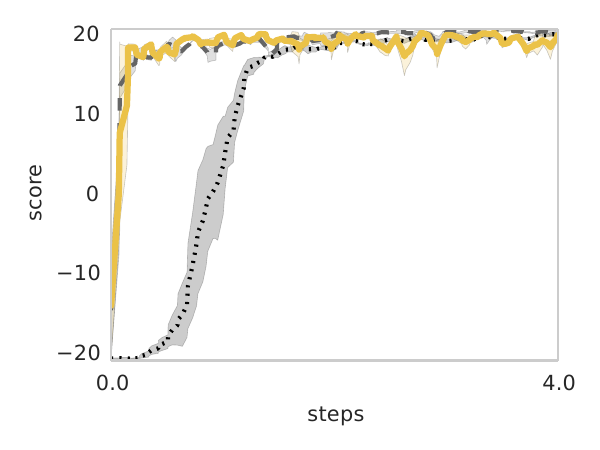} \\

	Target: H-flip & Target: HV-flip \\
        \includegraphics[width=.44\textwidth]{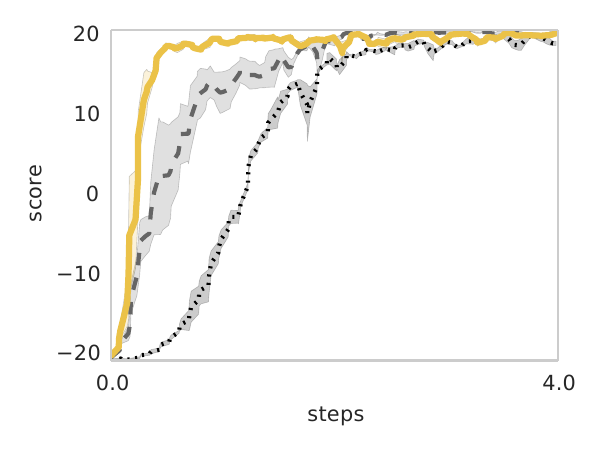} &
        \includegraphics[width=.44\textwidth]{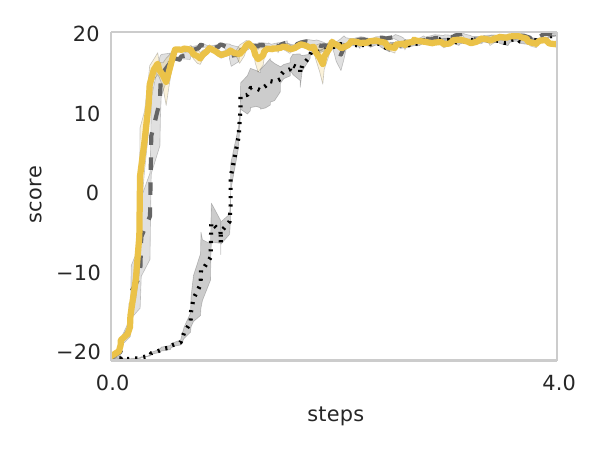} \\

	Target: Noisy & Target: V-flip \\
	\includegraphics[width=.44\textwidth]{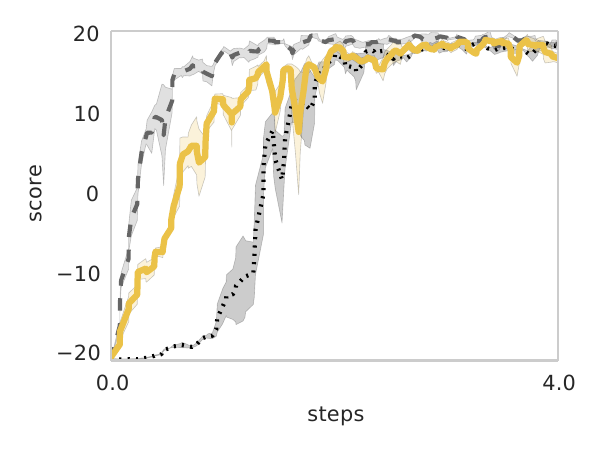} &
        \includegraphics[width=.44\textwidth]{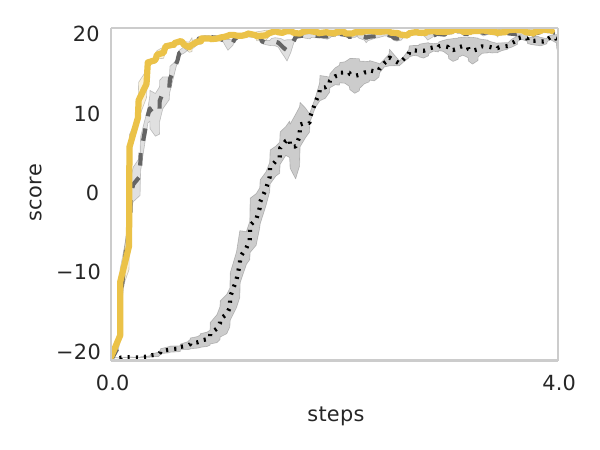} \\

	Target: White & Target: Zoom \\
        \includegraphics[width=.44\textwidth]{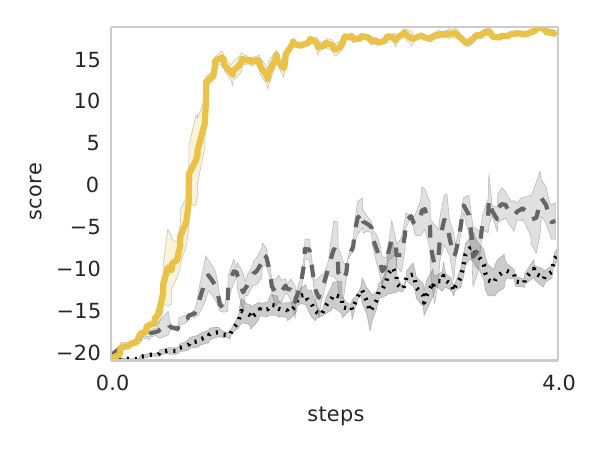} &
        \includegraphics[width=.44\textwidth]{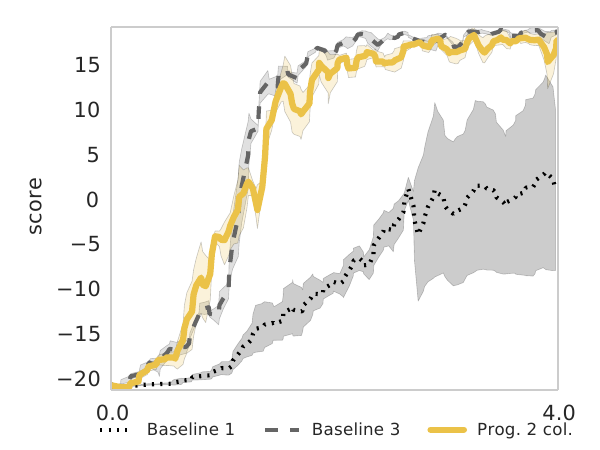} \\

    \end{tabular}
\caption{Training curves for transferring to 8 \textit{target} games after learning standard Pong first. }
    \label{fig:app_plot_pongs}
\end{figure}

\begin{figure}
     \begin{tabular}{cccccccc}
	Target: Track 1 & Target: Track 2 \\
        \includegraphics[width=.44\textwidth]{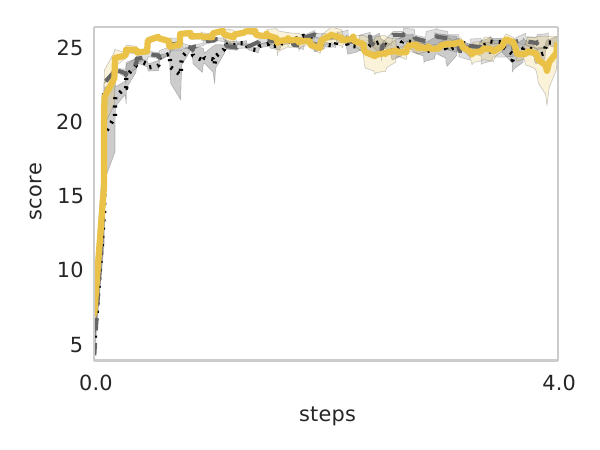} &
        \includegraphics[width=.44\textwidth]{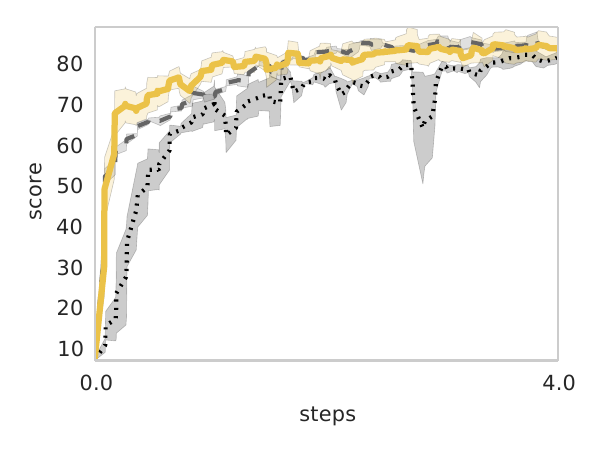} \\

	Target: Track 3 & Target: Track 4 \\
        \includegraphics[width=.44\textwidth]{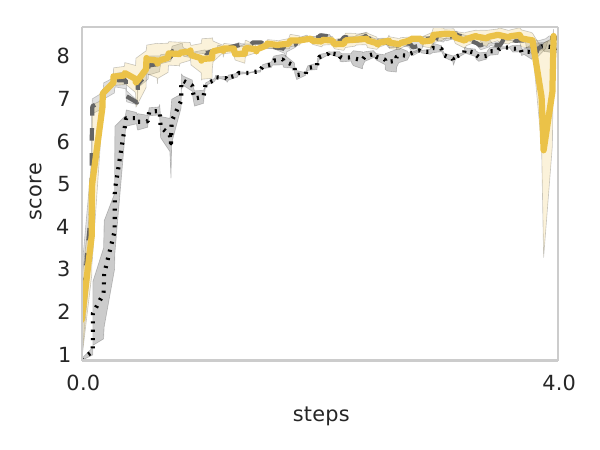} &
        \includegraphics[width=.44\textwidth]{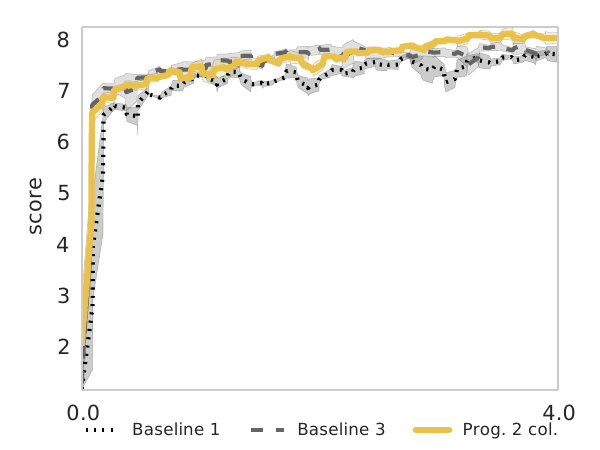} \\

	Target: Avoid 1 & Target: Avoid 2 \\
        \includegraphics[width=.44\textwidth]{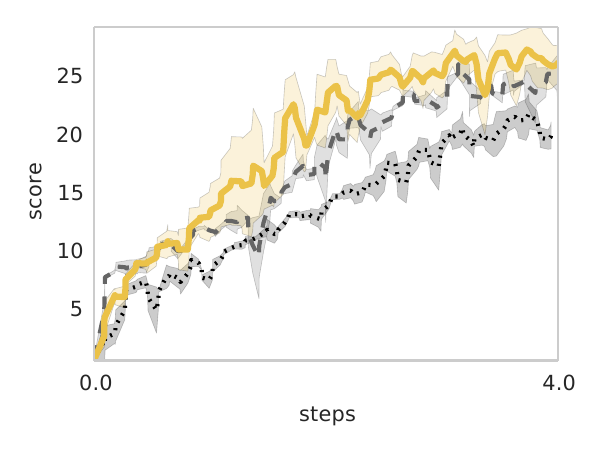} &
        \includegraphics[width=.44\textwidth]{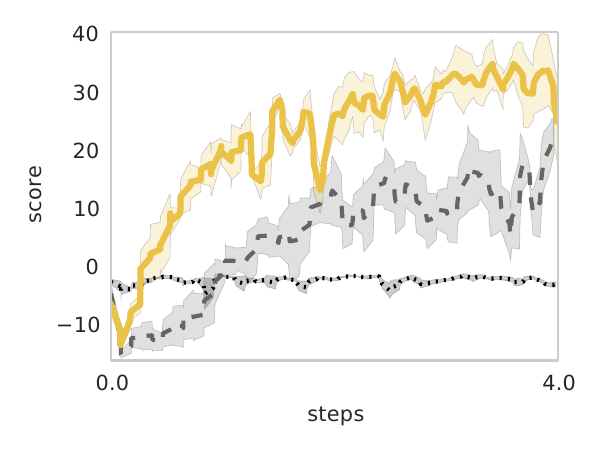} \\

	Target: Maze Y & Target: Maze M \\
        \includegraphics[width=.44\textwidth]{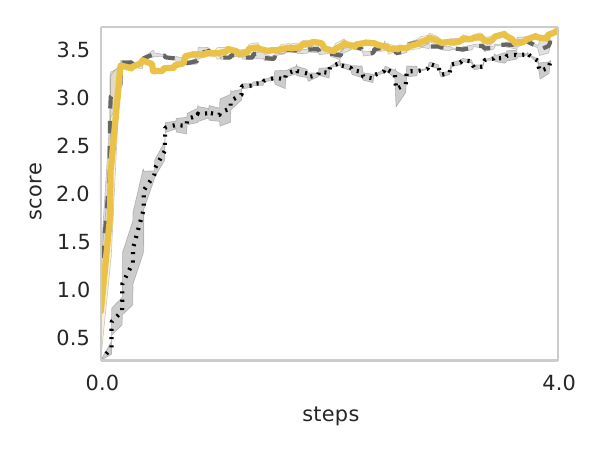} &
        \includegraphics[width=.44\textwidth]{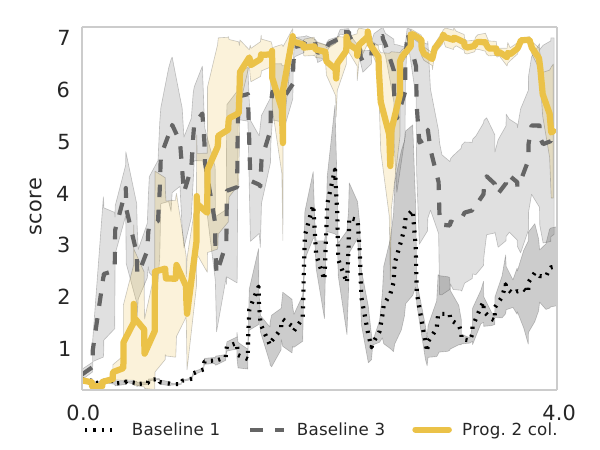} \\
    \end{tabular}
\caption{Training curves for transferring to 8 \textit{target} games after learning Maze Y first.}
    \label{fig:app_plot_lab}
\end{figure}

Figure \ref{fig:app_plot_pongs} shows how two-column progressive networks perform as compared to Baseline 3 (gray dashed line), a model pretrained on the \textit{source} game, here standard Pong, and then finetuned on a particular \textit{target} game, and Baseline 1 (black dotted line), where a single column is trained on standard Pong. Figure \ref{fig:app_plot_lab} shows two-column progressive networks and baselines on Labyrinth tasks; the \textit{source} game was Maze Y. 

%% file: app_lab.tex
\section{Labyrinth}

Section 5.4 evaluates progressive networks on foraging tasks in complex 3D maze
environments. Positive rewards are given to the agent for collecting apples and
strawberries, and negative rewards for mushrooms and lemons.
Episodes terminate when either all (positive) rewards are collected, or after a
fixed time interval.

Levels differ in their maze layout, the type of items present and the sparsity
of the reward structure. The levels we employed can be characterized as follows:
\begin{itemize}
\item Seek Track 1: simple corridor with many apples
\item Seek Track 2: U-shaped corridor with many strawberries
\item Seek Track 3: $\Omega$-shaped, with $90^o$ turns, with few apples
\item Seek Track 4: $\Omega$-shaped, with $45^o$ turns, with few apples
\item Seek Avoid 1: large square room with apples and lemons
\item Seek Avoid 2: large square room with apples and mushrooms
\item Seek Maze M : M-shaped maze, with apples at dead-ends
\item Seek Maze Y : Y-shaped maze, with apples at dead-ends
\end{itemize}